\documentclass{article}
\usepackage[preprint]{log_2026}			

\usepackage{booktabs}						
\usepackage{multirow}						
\usepackage{amsfonts}						
\usepackage{graphicx}						
\usepackage{wrapfig}
\usepackage{bm}
\usepackage{algorithm}
\usepackage{algpseudocode}
\usepackage{makecell}
\usepackage{adjustbox}	
\usepackage{tikz}
\usepackage{pgfplots}
\pgfplotsset{compat=1.18}
\usetikzlibrary{positioning,arrows.meta,fit,backgrounds}
\usepackage[numbers,compress,sort]{natbib}	

\usepackage{xcolor}

\title[Revisiting Simple Baselines for Relational Deep Learning]{Revisiting Simple Baselines for Relational Deep Learning}

\author{
    Viet The Nguyen$^{1}$,\hspace{5pt}
    Vijay Prakash Dwivedi$^{2,3}$,\hspace{5pt}
    Gunther Gust$^{1}$
    \\
    $^{1}$University of W\"urzburg, $^{2}$Seamless Labs, Inc., $^{3}$Stanford University
    \\
    \texttt{\{the-viet.nguyen,gunther.gust\}@uni-wuerzburg.de}
    \\
    \texttt{dwi@seamlesslabs.ai}
}

\begin{document}

\maketitle

\begin{abstract}
Relational deep learning has emerged as a powerful paradigm for learning end-to-end predictive models over relational databases by representing them as temporal heterogeneous graphs. Recent advances based on graph neural networks and Transformers have achieved strong performance across diverse domains. However, the effectiveness of simpler relational baselines remains underexplored, making it difficult to assess how much architectural complexity is needed to achieve strong predictive performance. Motivated by the observation that predictive signals arise from target attributes, relational attributes, and historical labels, we revisit how effectively these signals can be exploited using simple propagation methods. We adapt fixed low-pass feature propagation to relational entity graphs and train an MLP for prediction. The resulting predictions are further refined using Correct-and-Smooth, a graph-based label propagation algorithm that incorporates historical labels over task-specific meta-graphs connecting target entities. Across classification and regression tasks on RelBench, an MLP trained on neighborhood-filtered features outperforms one using only target-entity features on most tasks and recovers a substantial portion of the performance of heterogeneous GNNs. Correct-and-Smooth provides further gains from historical labels, although applying it requires a separately constructed meta-graph because connections among target entities are not directly specified by the relational entity graph. Together, our studies revisit the effectiveness of established propagation methods as simple RDL baselines while identifying canonical meta-graph construction as a key challenge for effective label propagation.
\end{abstract}

\section{Introduction}\label{sec:intro}
\begin{wrapfigure}{r}{0.45\textwidth}
	\vspace{-\intextsep}          
	\centering
	\includegraphics[width=0.45\textwidth]{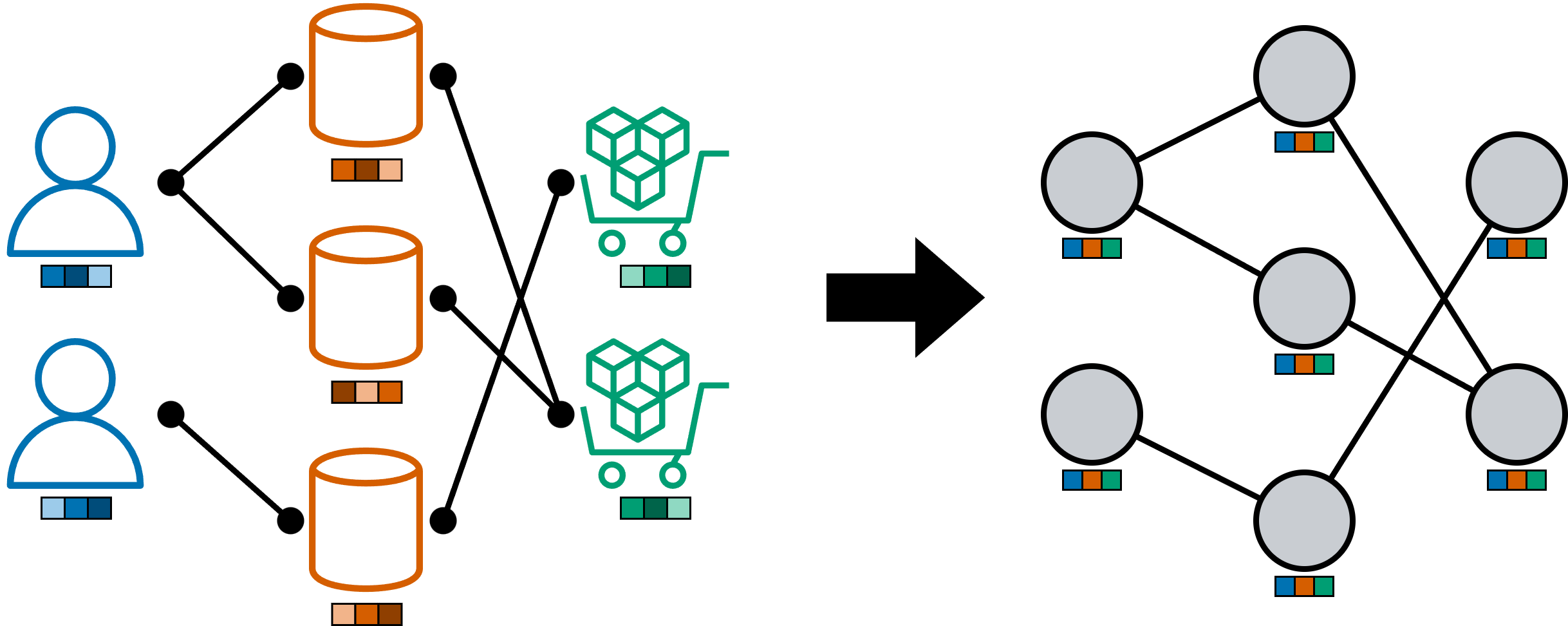}
	\caption{Example of graph conversion. (Left) A relational entity graph, where each node is an entity, and edges are primary--foreign key links. (Right) All nodes are collapsed into a single type, and feature sets are merged and normalized.}
	\label{fig:entity-graph}
	\vspace{-\intextsep}
\end{wrapfigure}
Data in domains such as e-commerce, healthcare, and social media is predominantly stored in relational databases, where tables are linked through primary--foreign key relationships. To reduce feature engineering efforts, relational deep learning (RDL) provides a framework for predictive modeling directly over such databases by representing each table row as a node in a temporal heterogeneous graph, whose edges encode primary--foreign key relationships~\cite{robinson2024relational,fey2024position}. This representation enables graph neural networks (GNNs) and transformer-based models to aggregate information across related entities, and has established RDL as an effective paradigm for learning on large-scale relational databases~\cite{wang2025griffin,chen2025relgnn,fey2025kumorfm,ranjan2026relational,hudovernik2026kumorfm,dwivedi2026relational}. 

Early approaches primarily adapted heterogeneous GNNs with relation-aware message passing to relational databases~\cite{fey2024position,chen2025relgnn,wang2025griffin}. Recent advances adopted transformer-based architectures with the aim of building relational foundation models, which achieved strong performance on large-scale benchmarks~\cite{dwivedi2026relational,ranjan2026relational}. In parallel, closed-source relational foundation models demonstrated that large-scale pretraining produces representations that transfer across databases and tasks~\cite{fey2025kumorfm,hudovernik2026kumorfm}. Across these models, the predictive signal on an entity-level task can be grouped into three sources: the attributes of the target entity, the attributes of entities related to it through foreign keys, and the historical labels of the target and of related entities. These models utilize target and relational attributes by sampling a temporal neighborhood around each target entity and aggregating its features through GNN message passing~\cite{fey2024position,chen2025relgnn} or Transformer attention~\cite{dwivedi2026relational,ranjan2026relational}. Recent work further shows that historical labels can add substantial predictive signals to improve zero-shot predictions~\cite{ranjan2026relational}.

Simple counterparts for exploiting these three sources are well established in graph learning. Specifically, an MLP operates directly on node features, fixed neighborhood propagation incorporates information from neighboring nodes without learned message passing~\cite{wu2019simplifying}, and Correct-and-Smooth (\emph{C\&S}) propagates label information over a graph~\cite{huang2021combining}. However, these methods are typically studied on homogeneous static graphs and cannot be applied to relational entity graphs directly. RDL graphs are heterogeneous, with nodes from different tables carrying different feature spaces. Additionally, they are temporal, requiring propagation to respect the time of each prediction, and labels are available only for the target entities of a task, rather than for all nodes in the graph. In particular, the graph induced by the database schema is not necessarily an appropriate graph for propagating task labels.

In this work, we make three contributions. First, we adapt established propagation methods to a simple RDL baseline that separates the three sources of predictive signal. We train an MLP on target attributes, then train the same MLP architecture on features enriched by fixed relational propagation~\cite{wu2019simplifying}, before refining its predictions with historical labels through the post-processing step with \emph{C\&S}~\cite{huang2021combining}. Second, we evaluate how much performance these stages recover on RelBench~\cite{robinson2024relbench,gu2026relbench}: feature propagation improves over the unfiltered MLP on 20 of 21 tasks and brings average classification AUROC within two points of the supervised heterogeneous GNN baseline. Finally, we study task-specific meta-graphs for label propagation and highlight canonical, temporally valid graph construction as a key challenge for exploiting historical labels in RDL.

\section{Simple Propagation for Relational Deep Learning} \label{sec:method}
We describe how feature and label propagation incorporate the three sources of predictive signal into the baseline. Appendices~\ref{app:feature-propagation} and~\ref{app:cs} give the feature construction, filter definition, and \emph{C\&S} updates.
\subsection{Relational Entity Graphs}
A relational database induces an entity graph $G=(\mathcal{V},\mathcal{E})$, where nodes are table rows and edges represent primary--foreign key links~\cite{fey2024position}. Each row carries attributes and, when applicable, a timestamp. An entity-level task provides labeled examples $(e,t)$, where $e$ is the target entity and $t$ is the seed time; predictions may use only information available by $t$. While RDL models typically learn relation-specific aggregation on this graph~\cite{fey2024position,chen2025relgnn}, we use it for fixed feature propagation before training an MLP.
\subsection{Feature Propagation}
\paragraph{Homogeneous relational graph} We combine all primary--foreign key links into one adjacency matrix and place each table's encoded attributes in a separate block of feature columns. This gives a homogeneous propagation graph while keeping attributes from different tables distinct (Figure~\ref{fig:entity-graph}). Appendix~\ref{app:feature-propagation} describes the encoding and normalization.
\paragraph{Low-pass propagation with temporal visibility} For each seed time $t$, we remove rows with later timestamps, retain static rows, and apply a fixed low-pass filter~\cite{wu2019simplifying} to the visible graph. Each propagation step combines a node's features with those of its neighbors using symmetric degree normalization. We then train an MLP on the resulting representations, using $K=2$ steps in the main experiments. The $K=0$ baseline uses target attributes alone, with structural features disabled. More details on the filtering process and feature definitions are also provided in Appendix~\ref{app:feature-propagation}.

\subsection{Label Propagation with Correct-and-Smooth}
After training on the filtered features, we incorporate historical labels through \emph{C\&S}~\cite{huang2021combining} to refine the MLP predictions. In particular, the \emph{Correct} stage propagates residuals from labeled nodes to adjust the MLP predictions, and the \emph{Smooth} stage propagates these predictions with observed labels as anchors. Since task labels belong to target entities that may be connected only indirectly in the relational graph, we construct a separate meta-graph over their prediction rows and historical labels.

We compare meta-graphs that attach the target's own history and related entities' labels directly or through temporal chains. Each construction admits only labels observed strictly before the seed time (Appendix~\ref{app:meta-graphs}) with propagation settings selected on validation before refining the MLP predictions.

\section{Experiments}
We evaluate whether the three simple propagation stages can provide a competitive relational baseline across the entity-level classification and regression tasks in RelBench-v1~\cite{robinson2024relbench} and RelBench-v2~\cite{gu2026relbench}. We exclude \texttt{rel-f1} and \texttt{rel-event}, for which temporal leakage has been reported~\cite{ranjan2026relational, xu2026no}. Following RelBench, classification is evaluated using AUROC~\cite{hanley1983method} and regression using MAE.

Our experiments follow the three stages introduced in Section~\ref{sec:method}. The \emph{unfiltered MLP} uses only the attributes of the target entity and corresponds to $K=0$. The \emph{filtered MLP} uses $K=2$ steps of fixed feature propagation and therefore incorporates attributes from related entities. Finally, \emph{filtered MLP + \emph{C\&S}} applies label propagation to the filtered predictions using the graph variant that incorporates both historical labels and neighboring labels (\textsc{Cross+Self}), which is described in Appendix~\ref{app:meta-graphs}. We select the \emph{C\&S} configuration on validation using the task metric (Appendix~\ref{app:cs}). All reported results are averaged over ten random seeds. Task statistics and training details are given in Appendices~\ref{app:datasets} and~\ref{app:details}. We compare our method against \emph{HeteroGNN}, the supervised RelBench baselines~\cite{robinson2024relbench,gu2026relbench}.

\subsection{Task Performance}
\begin{table}[t]
\centering\small\setlength{\tabcolsep}{2pt}
\caption{Entity classification on RelBench-v1 and RelBench-v2 (AUROC $\uparrow$).}
\label{tab:cls-results}
\scalebox{0.85}{%
\begin{tabular}{lcccc}
\toprule
\textbf{Task} & \makecell{\textbf{MLP (target attributes)}} & \makecell{ \textbf{MLP (+relational attributes)}} & \makecell{\textbf{C\&S (+historical labels)}} & \textbf{HeteroGNN} \\
\midrule
\multicolumn{5}{l}{\emph{RelBench-v1}} \\
\texttt{amazon}/\texttt{user-churn} & $53.65{\scriptstyle\pm0.18}$ & $68.12{\scriptstyle\pm0.05}$ & $\mathbf{68.69}{\scriptstyle\pm0.04}$ & \underline{$\mathbf{70.42}{\scriptstyle\pm0.05}$} \\
\texttt{amazon}/\texttt{item-churn} & $66.60{\scriptstyle\pm0.44}$ & $80.58{\scriptstyle\pm0.08}$ & $\mathbf{81.03}{\scriptstyle\pm0.07}$ & \underline{$\mathbf{82.81}{\scriptstyle\pm0.03}$} \\
\texttt{avito}/\texttt{user-clicks} & $51.74{\scriptstyle\pm0.51}$ & $65.08{\scriptstyle\pm1.06}$ & $\mathbf{65.10}{\scriptstyle\pm1.06}$ & \underline{$\mathbf{65.90}{\scriptstyle\pm1.95}$} \\
\texttt{avito}/\texttt{user-visits} & $49.64{\scriptstyle\pm0.01}$ & $\mathbf{66.29}{\scriptstyle\pm0.20}$ & \underline{$\mathbf{66.41}{\scriptstyle\pm0.18}$} & $66.20{\scriptstyle\pm0.10}$ \\
\texttt{hm}/\texttt{user-churn} & $55.25{\scriptstyle\pm0.05}$ & $68.21{\scriptstyle\pm0.06}$ & $\mathbf{69.16}{\scriptstyle\pm0.05}$ & \underline{$\mathbf{69.88}{\scriptstyle\pm0.21}$} \\
\texttt{stack}/\texttt{user-badge} & $51.93{\scriptstyle\pm1.87}$ & $86.04{\scriptstyle\pm0.08}$ & $\mathbf{86.16}{\scriptstyle\pm0.09}$ & \underline{$\mathbf{88.86}{\scriptstyle\pm0.08}$} \\
\texttt{stack}/\texttt{user-engagement} & $53.04{\scriptstyle\pm0.04}$ & $88.53{\scriptstyle\pm0.12}$ & $\mathbf{89.02}{\scriptstyle\pm0.09}$ & \underline{$\mathbf{90.59}{\scriptstyle\pm0.09}$} \\
\texttt{trial}/\texttt{study-outcome} & $65.77{\scriptstyle\pm0.36}$ & $66.37{\scriptstyle\pm0.34}$ & $\mathbf{67.68}{\scriptstyle\pm0.39}$ & \underline{$\mathbf{68.60}{\scriptstyle\pm1.01}$} \\ \midrule
\textbf{Average} & $55.95$ & $73.65$ & $\mathbf{74.16}$ & \underline{$\mathbf{75.41}$} \\
\midrule
\multicolumn{5}{l}{\emph{RelBench-v2}} \\
\texttt{ratebeer}/\texttt{user-churn} & $63.98{\scriptstyle\pm0.00}$ & $\mathbf{89.73}{\scriptstyle\pm0.12}$ & $\mathbf{89.73}{\scriptstyle\pm0.11}$ & \underline{$\mathbf{94.27}{\scriptstyle\pm0.22}$} \\
\texttt{ratebeer}/\texttt{beer-churn} & $62.40{\scriptstyle\pm0.84}$ & \underline{$\mathbf{78.93}{\scriptstyle\pm0.07}$} & $\mathbf{78.90}{\scriptstyle\pm0.08}$ & $78.67{\scriptstyle\pm0.60}$ \\
\texttt{ratebeer}/\texttt{brewer-dormant} & $63.67{\scriptstyle\pm0.32}$ & $79.73{\scriptstyle\pm0.04}$ & $\mathbf{79.78}{\scriptstyle\pm0.04}$ & \underline{$\mathbf{80.51}{\scriptstyle\pm0.17}$} \\
\texttt{arxiv}/\texttt{paper-citation} & $73.65{\scriptstyle\pm0.22}$ & $81.11{\scriptstyle\pm0.05}$ & $\mathbf{81.17}{\scriptstyle\pm0.05}$ & \underline{$\mathbf{82.50}{\scriptstyle\pm0.04}$} \\ \midrule
\textbf{Average} & $65.93$ & $82.38$ & $\mathbf{82.40}$ & \underline{$\mathbf{83.99}$} \\
\bottomrule
\end{tabular}}
\end{table}

\begin{table}[t]
\centering\small\setlength{\tabcolsep}{2pt}
\caption{Entity regression on RelBench-v1 and RelBench-v2 (MAE $\downarrow$).}
\label{tab:reg-results}
\scalebox{0.83}{%
\begin{tabular}{lcccc}
\toprule
\textbf{Task} & \makecell{\textbf{MLP (target attributes)}} & \makecell{ \textbf{MLP (+relational attributes)}} & \makecell{\textbf{C\&S (+historical labels)}} & \textbf{HeteroGNN} \\
\midrule
\multicolumn{5}{l}{\emph{RelBench-v1}} \\
\texttt{amazon}/\texttt{user-ltv} & $16.757{\scriptstyle\pm0.007}$ & $\mathbf{14.656}{\scriptstyle\pm0.013}$ & $\mathbf{14.656}{\scriptstyle\pm0.013}$ & \underline{$\mathbf{14.313}{\scriptstyle\pm0.013}$} \\
\texttt{amazon}/\texttt{item-ltv} & $59.958{\scriptstyle\pm0.150}$ & $\mathbf{50.994}{\scriptstyle\pm0.168}$ & $\mathbf{50.994}{\scriptstyle\pm0.168}$ & \underline{$\mathbf{50.053}{\scriptstyle\pm0.163}$} \\
\texttt{avito}/\texttt{ad-ctr} & $0.043{\scriptstyle\pm0.000}$ & \underline{$\mathbf{0.040}{\scriptstyle\pm0.003}$} & \underline{$\mathbf{0.040}{\scriptstyle\pm0.003}$} & $\mathbf{0.041}{\scriptstyle\pm0.001}$ \\
\texttt{hm}/\texttt{item-sales} & $0.076{\scriptstyle\pm0.000}$ & $\mathbf{0.060}{\scriptstyle\pm0.000}$ & $\mathbf{0.060}{\scriptstyle\pm0.000}$ & \underline{$\mathbf{0.056}{\scriptstyle\pm0.000}$} \\
\texttt{stack}/\texttt{post-votes} & $0.068{\scriptstyle\pm0.000}$ & $\mathbf{0.066}{\scriptstyle\pm0.000}$ & $\mathbf{0.066}{\scriptstyle\pm0.000}$ & \underline{$\mathbf{0.065}{\scriptstyle\pm0.000}$} \\
\texttt{trial}/\texttt{site-success} & $\mathbf{0.426}{\scriptstyle\pm0.002}$ & $0.437{\scriptstyle\pm0.006}$ & $0.443{\scriptstyle\pm0.006}$ & \underline{$\mathbf{0.400}{\scriptstyle\pm0.020}$} \\
\texttt{trial}/\texttt{study-adverse} & $48.203{\scriptstyle\pm0.676}$ & $44.867{\scriptstyle\pm0.151}$ & $\mathbf{44.817}{\scriptstyle\pm0.150}$ & \underline{$\mathbf{44.473}{\scriptstyle\pm0.209}$} \\
\midrule
\multicolumn{5}{l}{\emph{RelBench-v2}} \\
\texttt{ratebeer}/\texttt{user-count} & $14.453{\scriptstyle\pm0.006}$ & $\mathbf{9.115}{\scriptstyle\pm0.075}$ & $\mathbf{9.115}{\scriptstyle\pm0.075}$ & \underline{$\mathbf{7.374}{\scriptstyle\pm0.102}$} \\
\texttt{arxiv}/\texttt{author-publication} & $0.577{\scriptstyle\pm0.001}$ & \underline{$\mathbf{0.489}{\scriptstyle\pm0.002}$} & \underline{$\mathbf{0.489}{\scriptstyle\pm0.002}$} & $\mathbf{0.513}{\scriptstyle\pm0.008}$ \\
\bottomrule
\end{tabular}}
\end{table}

Tables~\ref{tab:cls-results} and~\ref{tab:reg-results} report the performance on the entity classification and entity regression tasks, respectively. The best result per task is bold and underlined; the second best is bold. For classification, feature propagation improves all twelve tasks, raising average AUROC from $55.95$ to $73.65$ on RelBench-v1 and from $65.93$ to $82.38$ on RelBench-v2 (Table~\ref{tab:cls-results}). As a result, the filtered MLP comes within $1.8$ and $1.6$ points of HeteroGNN, respectively, without learned relational aggregation. Further refinement with \emph{C\&S} adds about $0.51$ and $0.02$ points on average, with the largest gains on \texttt{trial}/\texttt{study-outcome} ($+1.31$) and \texttt{hm}/\texttt{user-churn} ($+0.95$).
For regression, feature propagation lowers MAE on eight of nine tasks and matches or exceeds HeteroGNN on \texttt{avito}/\texttt{ad-ctr} and \texttt{arxiv}/\texttt{author-publication} (Table~\ref{tab:reg-results}). The only exception to these feature-propagation gains is \texttt{trial}/\texttt{site-success}, where MAE rises from $0.426$ to $0.437$. By contrast, post-processing with \emph{C\&S} provides almost no regression improvement, which may reflect its original design for node classification~\cite{huang2021combining}. Consequently, validation tuning allows the propagation effect to be very small, and the selected configurations can largely preserve the filtered MLP's performance.
\subsection{Parameter Reduction}
\begin{figure}[t]
\centering
\begin{minipage}[c]{0.67\textwidth}
\centering
\includegraphics[width=\linewidth]{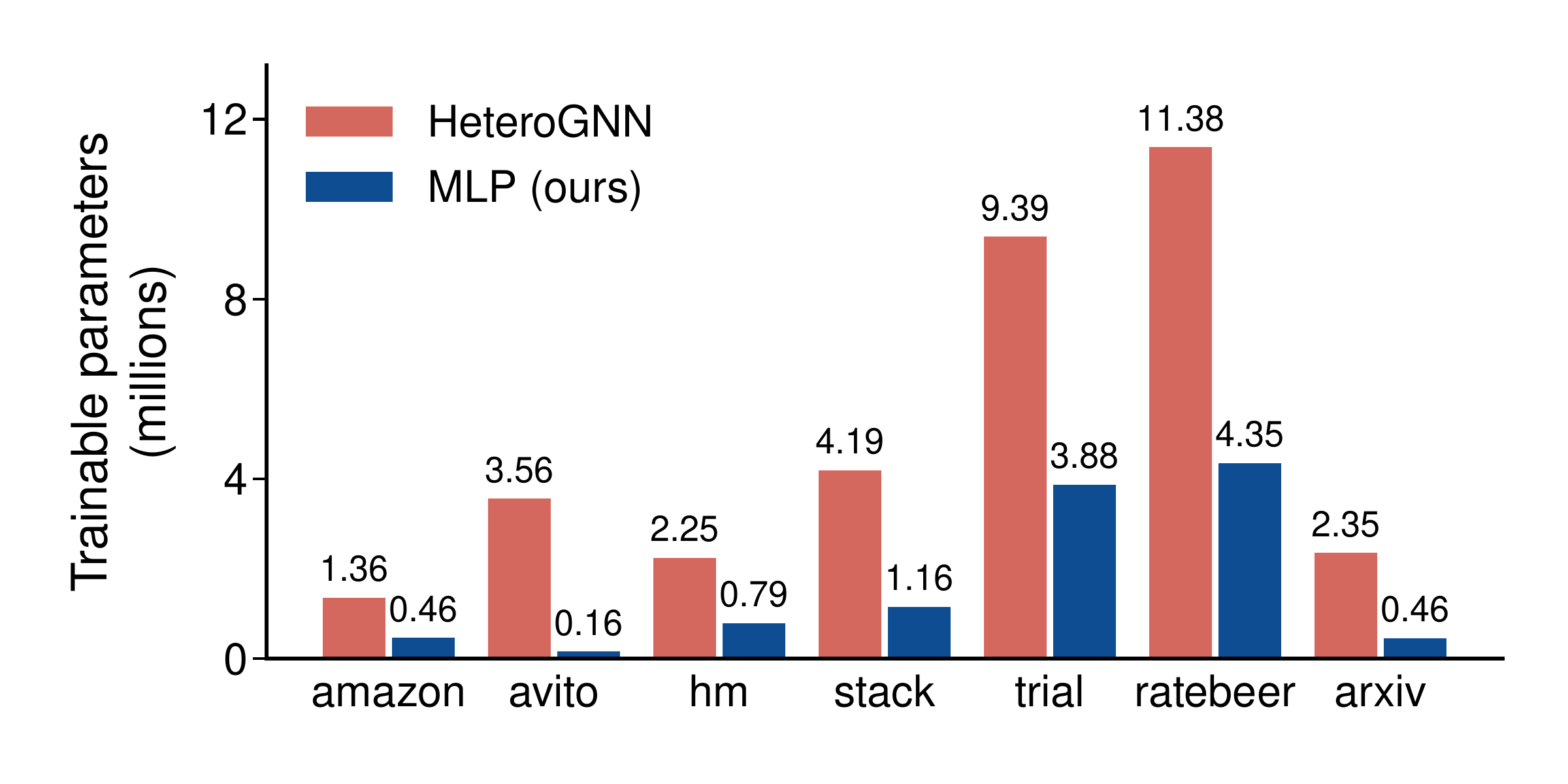}
\end{minipage}\hfill
\begin{minipage}[c]{0.29\textwidth}
\centering\small
\begin{tabular}{@{}lr@{}}
\toprule
\textbf{Database} & \textbf{Reduction} \\
\midrule
\texttt{rel-amazon} & $65.9\%$ \\
\texttt{rel-avito} & $95.5\%$ \\
\texttt{rel-hm} & $64.8\%$ \\
\texttt{rel-stack} & $72.3\%$ \\
\texttt{rel-trial} & $58.7\%$ \\
\texttt{rel-ratebeer} & $61.8\%$ \\
\texttt{rel-arxiv} & $80.4\%$ \\
\bottomrule
\end{tabular}

\end{minipage}
\caption{Trainable parameters (left) and MLP parameter reduction relative to HeteroGNN (right).}
\label{fig:params-compare}
\end{figure}
In addition, our method significantly reduces the number of trainable parameters by training only the MLPs, compared with HeteroGNN (Figure~\ref{fig:params-compare}), since feature propagation and \emph{C\&S} use only fixed operators. To examine whether these effects depend on different graph constructions, we perform label propagation on further variants in Appendix~\ref{app:csresults}.
\section{Discussion}
In conclusion, we evaluate an MLP on target attributes, enrich its inputs through fixed feature propagation, and refine its predictions with historical labels through \emph{C\&S}. Our results show that fixed feature propagation recovers substantial relational signal without learned aggregation, providing a simple baseline for assessing the gains from more complex RDL models. However, historical labels pose a different challenge: the schema connects related entities but does not specify which connections support label propagation. The variation across meta-graphs therefore motivates task-aware constructions that capture useful label dependencies while respecting temporal visibility alongside more efficient preprocessing to reduce the cost of filtering each seed-time snapshot (Appendix~\ref{app:filtercost}).





\bibliographystyle{unsrtnat}
\bibliography{reference}

\appendix
\section{Benchmark Tasks and Constructed Graphs}
\label{app:datasets}
This section describes how we build the homogeneous graph for feature filtering based on the database schema, and how the tasks in the datasets are defined with labels.

\paragraph{Databases.} Each database is converted once into the collapsed graph and stacked feature matrix described in Section~\ref{sec:method}, independently of the task evaluated on that database. Table~\ref{tab:app-graphs} summarizes the resulting representations, reporting the number of graph rows, the dimensionality of the feature matrix, the distribution of this dimensionality across text, numerical, and other column types, and the number of parameters in the resulting MLP. In particular, feature dimensions and model sizes vary substantially across databases. For example, \texttt{rel-avito} has $622$ input dimensions, while \texttt{rel-trial} has $15{,}145$, mainly from text, and an MLP roughly $24\times$ larger. This feature dimensionality, rather than the number of labeled examples, is the primary factor driving precomputation costs for some databases (Appendix~\ref{app:filtercost}).

\paragraph{Tasks.} Within each database, each task provides labeled examples, which RelBench splits into train, validation, and test sets according to the chronological order~\cite{robinson2024relbench}. Table~\ref{tab:app-tasks} reports the sizes of these splits together with the number of distinct seed times, which determines the number of filtered snapshots computed for each task. Tasks defined on the same database share the underlying graph but require separate snapshots. Consequently, tasks with more seed times incur higher precomputation costs. For example, \texttt{hm}/\texttt{user-churn}, with $54$ seed times, requires substantially more snapshot computation than the \texttt{avito} tasks, which have $5$ seed times. We evaluate all entity-level classification and regression tasks except those from \texttt{rel-event}, for which temporal leakage has been reported, and \texttt{rel-f1}, which \cite{xu2026no} identify as likely compromised. For schema diagrams and task semantics, we refer to the benchmark papers~\cite{robinson2024relbench,gu2026relbench}.

\begin{table}[ht]
\centering 
\caption{Task sizes by official split and number of distinct seed times.}
\label{tab:app-tasks}
\begin{tabular}{lrrrr}
\toprule
Task & Train & Val & Test & Seed times \\
\midrule
\multicolumn{5}{l}{\emph{RelBench-v1}} \\
\multicolumn{5}{l}{\emph{classification}} \\
\texttt{amazon}/\texttt{user-churn} & 4{,}708{,}383 & 409{,}792 & 351{,}885 & 33 \\
\texttt{amazon}/\texttt{item-churn} & 2{,}536{,}014 & 177{,}689 & 166{,}842 & 33 \\
\texttt{avito}/\texttt{user-clicks} & 59{,}454 & 21{,}183 & 47{,}996 & 5 \\
\texttt{avito}/\texttt{user-visits} & 86{,}619 & 29{,}979 & 36{,}129 & 5 \\
\texttt{hm}/\texttt{user-churn} & 3{,}832{,}692 & 76{,}556 & 74{,}575 & 54 \\
\texttt{stack}/\texttt{user-badge} & 3{,}386{,}276 & 247{,}398 & 255{,}360 & 42 \\
\texttt{stack}/\texttt{user-engagement} & 1{,}360{,}850 & 85{,}838 & 88{,}137 & 48 \\
\texttt{trial}/\texttt{study-outcome} & 11{,}994 & 960 & 825 & 21 \\
\midrule
\multicolumn{5}{l}{\emph{regression}} \\
\texttt{amazon}/\texttt{user-ltv} & 4{,}708{,}383 & 409{,}792 & 351{,}885 & 33 \\
\texttt{amazon}/\texttt{item-ltv} & 2{,}707{,}679 & 166{,}978 & 178{,}334 & 33 \\
\texttt{avito}/\texttt{ad-ctr} & 5{,}100 & 1{,}766 & 1{,}816 & 5 \\
\texttt{hm}/\texttt{item-sales} & 5{,}488{,}184 & 105{,}542 & 105{,}542 & 54 \\
\texttt{stack}/\texttt{post-votes} & 2{,}453{,}921 & 156{,}216 & 160{,}903 & 48 \\
\texttt{trial}/\texttt{site-success} & 151{,}407 & 19{,}740 & 22{,}617 & 21 \\
\texttt{trial}/\texttt{study-adverse} & 43{,}335 & 3{,}596 & 3{,}098 & 21 \\
\midrule
\multicolumn{5}{l}{\emph{RelBench-v2}} \\
\multicolumn{5}{l}{\emph{classification}} \\
\texttt{ratebeer}/\texttt{user-churn} & 373{,}709 & 19{,}908 & 9{,}392 & 76 \\
\texttt{ratebeer}/\texttt{beer-churn} & 2{,}470{,}686 & 92{,}367 & 79{,}927 & 76 \\
\texttt{ratebeer}/\texttt{brewer-dormant} & 98{,}697 & 15{,}840 & 16{,}366 & 20 \\
\texttt{arxiv}/\texttt{paper-citation} & 534{,}233 & 155{,}845 & 193{,}696 & 10 \\
\midrule
\multicolumn{5}{l}{\emph{regression}} \\
\texttt{ratebeer}/\texttt{user-count} & 373{,}709 & 19{,}908 & 9{,}392 & 76 \\
\texttt{arxiv}/\texttt{author-publication} & 210{,}769 & 39{,}015 & 39{,}655 & 10 \\
\bottomrule
\end{tabular}
\end{table}

\begin{table}[ht]
\centering \small 
\caption{Graph and feature statistics by database. Input dimensions include degree and age, used only for $K>0$.}
\label{tab:app-graphs}
\begin{tabular}{lrrrrrrr}
\toprule
Database & Tables & Nodes & Input dim & Text cols & Num.\ cols & Other & MLP params \\
\midrule
\multicolumn{8}{l}{\emph{RelBench-v1}} \\
\texttt{rel-amazon} & 3 & 15{,}000{,}713 & 1{,}814 & 6 & 7 & 2 & 464{,}897 \\
\texttt{rel-avito} & 8 & 20{,}679{,}117 & 622 & 2 & 40 & 0 & 159{,}745 \\
\texttt{rel-hm} & 3 & 16{,}664{,}809 & 3{,}092 & 10 & 13 & 14 & 792{,}065 \\
\texttt{rel-stack} & 7 & 4{,}247{,}264 & 4{,}529 & 15 & 29 & 6 & 1{,}159{,}937 \\
\texttt{rel-trial} & 15 & 5{,}434{,}924 & 15{,}145 & 50 & 53 & 30 & 3{,}877{,}633 \\
\midrule
\multicolumn{8}{l}{\emph{RelBench-v2}} \\
\texttt{rel-ratebeer} & 13 & 13{,}787{,}005 & 16{,}992 & 56 & 66 & 50 & 4{,}350{,}465 \\
\texttt{rel-arxiv} & 6 & 2{,}146{,}112 & 1{,}803 & 6 & 15 & 0 & 462{,}081 \\
\bottomrule
\end{tabular}
\end{table}

\section{Feature Propagation and Training Details}
\label{app:details}\label{app:feature-propagation}
In this section, we describe in detail how to encode and normalize attributes once per database, then append structural features for each visible snapshot.
\paragraph{Step 1: Column encoding.} We encode each table $T_i$ without learned parameters. Numerical columns use one scalar; text columns use $300$-dimensional GloVe sentence embeddings, as in RelBench. Categorical columns use indicators for the six most frequent values and the log-frequency of the row's value, counted through the validation timestamp. Other timestamp columns encode years relative to the row's time, or since 1990 for tables without a time column. Concatenating the encodings of all non-key columns of $T_i$ yields
$X_i \in \mathbb{R}^{n_i \times d_i}$, where $n_i = |T_i|$ and $d_i$ denotes the encoded dimensionality of table $T_i$. Table~\ref{tab:app-graphs} reports the resulting dimensions aggregated per database. Primary and foreign keys are excluded from the feature representation and are used solely to define the graph structure.

\paragraph{Step 2: Stacking.} For $m$ tables, we place the feature matrices in disjoint row and column blocks,
\begin{equation}
X = \operatorname{diag}(X_1,\ldots,X_m) \in \mathbb{R}^{n\times d},
\qquad n=\sum_{i=1}^{m} n_i, \qquad d=\sum_{i=1}^{m} d_i,
\label{eq:union}
\end{equation}
so each row contains its entity's encoded attributes in the block for its source table and zeros elsewhere.

\paragraph{Step 3: Row normalization.} SGC divides each feature vector by its signed row sum, which suits non-negative bag-of-words features but can become unstable for signed GloVe coordinates. For row $x_v$ of $X$, we instead use
\begin{equation}
\tilde{x}_v =
\begin{cases}
x_v/\lVert x_v\rVert_1, & \lVert x_v\rVert_1>0,\\
0, & \lVert x_v\rVert_1=0,
\end{cases}
\qquad \lVert x_v\rVert_1=\sum_{c=1}^{d}|X_{v,c}|.
\label{eq:absnorm}
\end{equation}
For nonzero, non-negative features, this agrees with signed-sum normalization; zero rows remain zero, and the empty column blocks do not affect the norm. Consequently, omitting this normalization and the symmetric adjacency reduced test AUROC by up to $41$ percentage points.

\textbf{Step 4: Structural columns.} After normalization, we append two structural features,
\begin{equation}
\bar{x}_v(t) \;=\; \bigl[\; \tilde{x}_v \;\big|\; \log\!\left(1 + \deg_{A_t}(v)\right) \;\big|\; a_v(t) \;\bigr]
\;\in\; \mathbb{R}^{d+2},
\label{eq:structcols}
\end{equation}
where $\bm{A}_t$ contains only rows visible at $t$ and $\deg_{A_t}(v)$ excludes self-loops. The age $a_v(t)=\min(t-t_v,50)$ is measured in years and set to zero for rows without timestamps. We recompute both features for each snapshot and append them after row normalization to preserve their absolute scale. Stacking $\bar{x}_v(t)$ gives $\bar{\bm{X}}_t$, which the filter propagates together with the encoded attributes.

\subsection{Filter Definition}
\label{app:filterdef}
For the entity graph $G=(\mathcal{V},\mathcal{E})$ of Section~\ref{sec:method}, let $\bm{A}_t$ be the symmetric adjacency induced by the visible rows:
\begin{equation}
\mathcal{V}_t = \{v\in\mathcal{V}: v\text{ has no timestamp or }t_v\leq t\}.
\end{equation}
Adding self-loops gives $\tilde{\bm{A}}_t=\bm{A}_t+\bm{I}$, with degree matrix $\tilde{\bm{D}}_t$. The normalized filter and its output are defined as:
\begin{equation}
\begin{aligned}
\bm{S}_t&=\tilde{\bm{D}}_t^{-1/2}\tilde{\bm{A}}_t\tilde{\bm{D}}_t^{-1/2}, \\
\ \bm{H}_t&=\bm{S}_t^K\bar{\bm{X}}_t,
\label{eq:filter}
\end{aligned}
\end{equation}
where $\bar{\bm{X}}_t$ contains the normalized attributes and snapshot-specific structural features from Eq.~\eqref{eq:structcols}. The MLP then takes $h_{e,t}=(\bm{H}_t)_e$ for prediction row $(e,t)$. Setting $K=0$ and both structural columns to zero gives the target-attribute baseline.

\subsection{Graph Filtering Cost}
\label{app:filtercost}
The filter is precomputed per task: the collapsed graph and the stacked features are shared by
all tasks of a database, but one snapshot is computed per distinct seed time of the task and
read out at the task's entity rows, so cost grows with the seed-time count of
Table~\ref{tab:app-tasks}. We measured the filtering time, ranging from $5.5$ minutes (\texttt{rel-avito}) to $47$ minutes
(\texttt{rel-stack}) for one representative task per database; on \texttt{rel-avito}, deepening
the filter to $K=10$ grows this to $113$ minutes. Dense design matrices dominate memory, requiring roughly $(\text{rows}) \times (d+2) \times 4$ bytes per snapshot. Serial peak memory ranged from $44$\,GB (\texttt{rel-avito}) to $214$\,GB (\texttt{rel-amazon}). This process can be further sped up by utilizing GPUs for filtering, which remains as future work.

\paragraph{Parallel filtering and timing.} To reduce the filtering time, we process independent snapshots and feature columns across 32 parallel jobs, producing the same features as serial filtering. Table~\ref{tab:app-parallel-filter} reports filtering wall time with all column types at $K=2$, MLP training time, and HeteroGNN training time of one run~\cite{robinson2024relbench,gu2026relbench}. We note that the timings exclude column encoding cached for both methods and include HeteroGNN's neighborhood sampling during training.

\begin{table}[ht]
\centering\small\setlength{\tabcolsep}{4pt}
\caption{Filtering and training time per task (minutes); timing protocol in Appendix~\ref{app:filtercost}.}
\label{tab:app-parallel-filter}
\scalebox{0.95}{%
\begin{tabular}{lcc|c}
\toprule
Task & Filtering & Training MLP & HeteroGNN \\
\midrule
\multicolumn{4}{l}{\emph{RelBench-v1}} \\
\multicolumn{4}{l}{\emph{classification}} \\
\texttt{amazon}/\texttt{user-churn} & $8.5$ & $16.2$ & $28.8$ \\
\texttt{amazon}/\texttt{item-churn} & $9.2$ & $10.2$ & $15.3$ \\
\texttt{avito}/\texttt{user-clicks} & $1.1$ & $0.3$ & $9.6$ \\
\texttt{avito}/\texttt{user-visits} & $1.8$ & $0.3$ & $3.9$ \\
\texttt{hm}/\texttt{user-churn} & $1.1$ & $25.3$ & $16.3$ \\
\texttt{stack}/\texttt{user-badge} & $1.4$ & $18.7$ & $21.9$ \\
\texttt{stack}/\texttt{user-engagement} & $1.4$ & $14.2$ & $30.3$ \\
\texttt{trial}/\texttt{study-outcome} & $2.0$ & $1.0$ & $4.9$ \\
\midrule
\multicolumn{4}{l}{\emph{regression}} \\
\texttt{amazon}/\texttt{user-ltv} & $11.5$ & $23.0$ & $16.4$ \\
\texttt{amazon}/\texttt{item-ltv} & $13.6$ & $18.8$ & $13.6$ \\
\texttt{avito}/\texttt{ad-ctr} & $3.1$ & $0.05$ & $1.9$ \\
\texttt{hm}/\texttt{item-sales} & $1.2$ & $35.3$ & $16.4$ \\
\texttt{stack}/\texttt{post-votes} & $1.6$ & $26.7$ & $25.4$ \\
\texttt{trial}/\texttt{site-success} & $2.0$ & $5.8$ & $4.1$ \\
\texttt{trial}/\texttt{study-adverse} & $1.8$ & $3.4$ & $1.8$ \\
\midrule
\multicolumn{4}{l}{\emph{RelBench-v2}} \\
\multicolumn{4}{l}{\emph{classification}} \\
\texttt{ratebeer}/\texttt{user-churn} & $11.2$ & $13.4$ & $15.7$ \\
\texttt{ratebeer}/\texttt{beer-churn} & $31.2$ & $89.9$ & $46.2$ \\
\texttt{ratebeer}/\texttt{brewer-dormant} & $25.7$ & $5.6$ & $11.1$ \\
\texttt{arxiv}/\texttt{paper-citation} & $1.5$ & $3.0$ & $13.8$ \\
\midrule
\multicolumn{4}{l}{\emph{regression}} \\
\texttt{ratebeer}/\texttt{user-count} & $4.5$ & $12.6$ & $16.3$ \\
\texttt{arxiv}/\texttt{author-publication} & $0.9$ & $1.2$ & $3.4$ \\
\bottomrule
\end{tabular}}
\end{table}

\subsection{Training Details}
\label{app:training}
The prediction head follows the reference MLP of~\cite{huang2021combining}, with batch normalization removed. It consists of a single hidden layer of width $256$, followed by ReLU activation and dropout with rate $0.5$. The same architecture is used across all datasets and is trained with Adam using a batch size of $512$, following RelBench configurations~\cite{robinson2024relbench}. We remove batch normalization because its running statistics are estimated from the training period and can degrade substantially under RelBench's temporal test shift. On one task, for example, removing batch normalization left validation AUROC unchanged but prevented test AUROC from collapsing from $0.886$ to $0.224$. Then, we train with binary cross-entropy for classification and $\ell_1$ loss for regression, clipping regression predictions to the 2nd--98th percentile range of training targets~\cite{robinson2024relbench}. Evaluation uses the official RelBench AUROC and MAE implementations, with MAE reported in the target's original units. We select the learning rate on validation from $\{10^{-3}, 3{\cdot}10^{-3}, 10^{-2}\}$ and stop after 5 epochs without improvement in the validation metric.

We use the official RelBench splits~\cite{robinson2024relbench,gu2026relbench} without modification. The benchmark fixes validation and test timestamps per database, generates labeled examples at a calendar stride, and masks test labels; we do not perform any additional re-splitting. All results are averaged over ten fixed seeds. To reproduce our results, readers can select the following seeds:
$\{0,7,13,21,42,77,123,555,1234,2026\}$,
which control MLP initialization and batch shuffling. The filter is deterministic on the relational entity graph. The baselines additionally employ per-edge-type message functions and column encoders on top of comparable input encoders, whereas our model uses a single prediction head after feature propagation. Thus, our prediction head has fewer trainable parameters by construction.

\paragraph{Comparison to the RDL training pipeline.} RelBench-style models such as RelGNN and RelGT sample a temporal neighborhood for each labeled row, restrict it to information visible at the seed time, and jointly train the relational model and prediction head. Our pipeline instead filters the full visible graph once per seed time, then trains an MLP on the resulting rows. This moves relational computation into preprocessing, allowing the filtered features to be computed once and reused.

\paragraph{Reproducibility.} Our code is available at \href{https://github.com/vietngth/revisit-baselines-rdl}{https://github.com/vietngth/revisit-baselines-rdl}.

\section{Label Propagation with Correct-and-Smooth}
\label{app:cs}
Correct-and-Smooth (\emph{C\&S})~\cite{huang2021combining} post-processes the predictions of a base
model with two label-propagation stages over a graph concerning the task entities. Let $Z$ denote the
base predictions, $Y$ the observed labels, and $L$ the labeled nodes of the graph. For weighted label-graph adjacency $A$ and degree matrix $D$, the two stages use normalizations $S_1$ and $S_2$ selected from $\{D^{-1/2} A D^{-1/2},\, D^{-1} A,\, A D^{-1}\}$. \emph{Correct} initializes residuals as $E_L = Y_L - Z_L$, with zeros elsewhere, and computes the following:
\begin{equation}
\begin{aligned}
E^{(0)} &= E, \\
E^{(k+1)} &= (1-\alpha_1)\, E + \alpha_1 S_1 E^{(k)}, \\
Z^{(r)} &= Z + s \odot E^{(50)},
\label{eq:correct}
\end{aligned}
\end{equation}
with three scaling options: none ($s=1$), FDiff-scale, a constant $s\in[1,50]$ with labeled residuals reset to $E_L$ each iteration, or Autoscale, for classification only, with $s_i=\sigma/|E_i^{(50)}|$, where $\sigma$ is the mean absolute labeled residual in the prediction row's graph. Next, \emph{Smooth} sets $G_L=Y_L$ and $G=Z^{(r)}$ elsewhere, then computes
\begin{equation}
\begin{aligned}
G^{(0)} &= G, \\
G^{(k+1)} &= (1-\alpha_2)\, G + \alpha_2 S_2 G^{(k)}.
\label{eq:smooth}
\end{aligned}
\end{equation}
Here, the algorithm is fixed with $k=50$ iterations, clipping to the training-label range after each iteration. Finally, we read out the seed prediction; seeds without labeled neighbors keep their base predictions. Both stages assume positive correlation of errors and labels along edges. We tune
$(\alpha_1, \alpha_2) \in [0,1]^2$, the normalizations $S_1$ and $S_2$, the residual scaling, and stage selection on validation using the task metric. We use a tree-structured Parzen estimator with $60$ trials for RelBench-v1 classification and $300$ otherwise, and apply the selected
configuration to all ten seeds. Setting both coefficients to zero recovers the base exactly,
which the search may select.

\subsection{Graph Constructions}
\label{app:csgraphs}\label{app:meta-graphs}
\emph{C\&S} requires a graph whose nodes correspond to the task's prediction rows and whose edges carry label information, but the relational schema does not provide such a graph directly. We therefore construct one for each prediction row $(e,t)$ using two sources of label information observable at time $t$: the entity's own historical labels and the historical labels of \emph{related entities}. Specifically, we identify related entities by composing the schema's foreign-key relations along walks that eventually return to the task's entity table. Two entities are connected with a weight proportional to the number of records they share, where each shared record contributes inversely to the number of entities that it connects, following the co-occurrence weighting scheme used in collaboration networks~\cite{newman2001scientific}. For each seed time $t$, $\mathcal{N}(e,t)$ contains up to $100$ related entities with the largest weights from shared records observed before $t$. Table~\ref{tab:app-graphnames} specifies the label sources and history limits, Figure~\ref{fig:csgraphs} shows their connections, and Algorithm~\ref{alg:cs} gives the shared construction steps.

\begin{algorithm}[ht]
\caption{Build a label graph for prediction row $(e,t)$}
\label{alg:cs}
\begin{algorithmic}[1]
\Require histories $\mathcal{H}$; related entities $\mathcal{N}(e,t)$ and weights $w_u$; base predictions $Z$; variant from Table~\ref{tab:app-graphnames}
\State Initialize an unlabeled seed $s$ with $Z_s=Z_{e,t}$; set $w_e=1$
\ForAll{$u\in\{e\}\cup\mathcal{N}(e,t)$}
  \State Read attachment type and cap $c$ from Table~\ref{tab:app-graphnames} ($c=0$ for ``None'')
  \State $L_u\gets\{(y,t_y)\in\mathcal{H}(u):t_y<t\}$, ordered newest first
  \State Skip $u$ if its attachment type is ``None'' or $L_u$ is empty
  \State $T_u\gets$ the first $\min(c,|L_u|)$ labels of $L_u$
  \If{labels attach directly} \label{line:star}
    \State Connect $s$ to each $y_i\in T_u$ with weight $0.7^{i-1}$ for \textsc{Self}, otherwise $w_u$
  \ElsIf{labels form a chain} \label{line:chain}
    \State Add the path $s-y_1-\cdots-y_{|T_u|}$; first edge weight $w_u$, remaining weights $1$
  \Else \Comment{relay}
    \State Create one relay per label in $L_u$ if $u=e$; otherwise one for the newest label
    \State Initialize each relay with its associated past label's base prediction
    \State Connect $s$ to each relay with weight $w_u$; attach a separate copy of $T_u$ with unit edges
  \EndIf
\EndFor
\State Initialize history nodes with their associated base predictions and observed labels
\State \Return the weighted undirected graph and seed $s$
\end{algorithmic}
\end{algorithm}

\begin{figure}[p]
\centering
\newcommand{\lgSeed}{\node[seed] (e) at (0,0) {$e$};}
\newcommand{\lgOwnStar}{%
  \foreach \i/\y in {1/0.65,2/0,3/-0.65} {
    \node[own] (a\i) at (-1.15,\y) {$y_\i$};
    \draw[edge] (e) -- (a\i);}}
\newcommand{\lgOwnChain}{%
  \foreach \i in {1,2,3} {\node[own] (a\i) at (-0.8*\i,0) {$y_\i$};}
  \draw[edge] (e) -- (a1) -- (a2) -- (a3);}
\newcommand{\lgCross}{%
  \foreach \j/\y in {1/0.5,2/-0.5} {
    \node[hist] (b\j) at (1.2,\y) {$y_{\j,1}$};
    \draw[edge] (e) -- (b\j);}}
\newcommand{\lgRelays}{%
  \foreach \j/\y in {1/0.9,2/-0.9} {
    \node[relay] (r\j) at (0.95,\y) {$v_\j$};
    \draw[edge] (e) -- (r\j);
    \foreach \k/\dy in {1/0.6,2/0,3/-0.6} {
      \node[hist] (b\j\k) at (2.05,\y+\dy) {$y_{\j,\k}$};
      \draw[edge] (r\j) -- (b\j\k);}}}
\newcommand{\lgOwnRelays}{%
  \foreach \i/\y in {1/0.9,m/-0.9} {
    \node[relay] (o\i) at (-0.95,\y) {$r_\i$};
    \draw[edge] (e) -- (o\i);
    \foreach \k/\dy in {1/0.6,2/0,3/-0.6} {
      \node[own] (a\i\k) at (-2.05,\y+\dy) {$y_\k$};
      \draw[edge] (o\i) -- (a\i\k);}}
  \node at (-0.95,0) {$\vdots$};}
\newcommand{\lgCrossChains}{%
  \foreach \j/\y in {1/0.5,2/-0.5} {
    \foreach \k in {1,2,3} {\node[hist] (b\j\k) at (0.8*\k,\y) {$y_{\j,\k}$};}
    \draw[edge] (e) -- (b\j1) -- (b\j2) -- (b\j3);}}
\newcommand{\lgContinue}{%
  \node[font=\small] (old) at (-3.0,0) {$\cdots$};
  \draw[edge] (a3) -- (old);
  \foreach \j/\y in {1/0.5,2/-0.5} {
    \node[font=\small] (old\j) at (3.0,\y) {$\cdots$};
    \draw[edge] (b\j3) -- (old\j);}}
\newcommand{\lgPanel}[5]{%
  \begin{scope}[shift={(#1,#2)}]
    \lgSeed #3
    \node[paneltitle] at (0,#5) {#4};
  \end{scope}}
\begin{tikzpicture}[
  every node/.style={font=\scriptsize},
  own/.style={circle,draw=black,fill=white,minimum size=5.5mm,inner sep=0pt},
  seed/.style={own,draw=green!40!black,very thick,fill=green!12},
  hist/.style={own,fill=black!15},
  relay/.style={own,densely dashed},
  edge/.style={draw=black!75,line width=0.5pt},
  paneltitle/.style={anchor=north,font=\small},
  guide/.style={font=\scriptsize,align=center}]
\node[seed] at (-1.5,1.4) {$e$};
\node[guide,anchor=west] at (-1.15,1.4) {prediction row};
\node[own] at (1.8,1.4) {$y_i$};
\node[guide,anchor=west] at (2.15,1.4) {own label};
\node[hist] at (4.6,1.4) {$y_{j,k}$};
\node[guide,anchor=west] at (4.95,1.4) {related label};
\node[relay] at (8.0,1.4) {$v_j$};
\node[guide,anchor=west] at (8.35,1.4) {unlabeled relay};
\node[guide,font=\small\bfseries] at (0,-0.05) {Direct and relay connections};
\node[guide,font=\small\bfseries] at (7,-0.05) {Temporal chains};
\lgPanel{0}{-1.7}{\lgOwnStar}{(a) \textsc{Self}}{-1.2}
\lgPanel{7}{-1.7}{\lgOwnChain}{(f) \textsc{Self-Chain}}{-1.2}
\lgPanel{0}{-4.3}{\lgCross}{(b) \textsc{Cross}}{-1.2}
\lgPanel{7}{-4.3}{\lgOwnChain\lgCross}{(g) \textsc{Chain+Cross}}{-1.2}
\lgPanel{0}{-6.9}{\lgOwnStar\lgCross}{(c) \textsc{Cross+Self}}{-1.2}
\lgPanel{7}{-6.9}{\lgOwnChain\lgCrossChains}{(h) \textsc{Chain+Hist-3}}{-1.2}
\lgPanel{0}{-10.5}{\lgRelays}{(d) \textsc{Cross+Hist}}{-2.0}
\lgPanel{7}{-10.5}{\lgOwnChain\lgCrossChains\lgContinue}{(i) \textsc{Chain+Hist}}{-2.0}
\path (current bounding box.center);
\pgfgetlastxy{\lgMidX}{\lgMidY}
\lgPanel{\lgMidX}{-14.8}{\lgOwnRelays\lgRelays}{(e) \textsc{All}}{-2.0}
\end{tikzpicture}
\caption{Meta-graphs for $(e,t)$ with undirected edges and labels observed before $t$. Own histories lie left of $e$, related histories right; indices increase from newest to oldest, with $y_{j,k}$ denoting related entity $v_j$'s $k$th label. Dashed relays are unlabeled intermediate nodes that carry base predictions and connect historical labels to the seed in two hops. In (e), $r_1,\ldots,r_m$ represent all $m$ own past labels, each with separate copies of up to three newest own labels. Table~\ref{tab:app-graphnames} gives the history limits.}
\label{fig:csgraphs}
\end{figure}

\begin{table}[ht]
\centering\small\setlength{\tabcolsep}{5pt}
\caption{Meta-graph variants in Figure~\ref{fig:csgraphs}. A star attaches labels directly; a chain orders them newest first; a \textbf{relay} is an unlabeled intermediate node between the seed and historical labels.}
\label{tab:app-graphnames}
\begin{tabular}{@{}llp{3.8cm}p{4.4cm}@{}}
\toprule
Variant & Panel & Own history & Related history (per entity) \\
\midrule
\textsc{Self} & (a) & Star, all labels & None \\
\textsc{Cross} & (b) & None & Most recent label, directly \\
\textsc{Cross+Self} & (c) & Star, all labels & Most recent label, directly \\
\textsc{Cross+Hist} & (d) & None & Relay, up to 3 labels \\
\textsc{All} & (e) & One relay per past label; up to 3 label copies each & Relay, up to 3 labels \\
\midrule
\textsc{Self-Chain} & (f) & Chain, all labels & None \\
\textsc{Chain+Cross} & (g) & Chain, all labels & Most recent label, directly \\
\textsc{Chain+Hist-3} & (h) & Chain, up to 3 & Chain, up to 3 labels \\
\textsc{Chain+Hist} & (i) & Chain, all labels & Chain, all labels \\
\bottomrule
\end{tabular}
\end{table}

All variants exclude labels observed at or after the seed time; boundary audits verify that adding such labels leaves the admitted history unchanged. With this common temporal rule, the variants differ in which histories they use and how they connect them to the prediction row. The following descriptions follow panels (a)--(i) in Figure~\ref{fig:csgraphs} and complement Table~\ref{tab:app-graphnames}.
\begin{enumerate}
\renewcommand{\labelenumi}{(\alph{enumi})}
\item \textbf{\textsc{Self}.} We connect the prediction row directly to all of the target entity's past labels. Each label is one hop away, but its edge weight decreases with age rank. For example, the newest label receives weight $1$, the next $0.7$, and so on. This construction uses the entity's own past outcomes as context, with greater edge weight on recent observations.

\item \textbf{\textsc{Cross}.} We instead use the most recent label of each selected related entity, connecting it directly to the prediction row with its co-occurrence weight. The target's own labels are omitted. This construction asks whether related entities provide useful label information, including when the target has no label history of its own.

\item \textbf{\textsc{Cross+Self}.} We attach both the target's full label history and the newest label of each selected related entity directly to the prediction row. Both sources are therefore one hop away. Unlike \textsc{Self}, own-label edges here have weight $1$ throughout, while related-label edges retain their co-occurrence weights. This is the construction used in the main results.

\item \textbf{\textsc{Cross+Hist}.} For each selected related entity, we insert an unlabeled relay between the prediction row and up to three of that entity's newest labels. The relay includes the base prediction associated with its newest label, while its attached history nodes carry the observed labels. This gives each related entity a local history around an intermediate prediction node; label information reaches the target through two hops. The target's own history is omitted.

\item \textbf{\textsc{All}.} We extend \textsc{Cross+Hist} with the target's own history, also through relays. Each own past label creates a separate unlabeled relay carrying the base prediction associated with that label. Every own relay receives fresh copies of the same up to three newest own labels, and all edges in these own-history branches have weight $1$. Thus, the number of own relays follows the full history length, while each relay's labeled neighborhood remains limited to three observations.

\item \textbf{\textsc{Self-Chain}.} We arrange the target's full history along a path: the prediction row connects to the newest label, that label to the next newest, and so on. All edges have weight $1$, and related entities are omitted. Older observations lie progressively more hops from the prediction row, so the topology encodes recency instead of assigning a different edge weight to each past label.

\item \textbf{\textsc{Chain+Cross}.} We retain the own-history path of \textsc{Self-Chain} and add a direct edge to the newest label of each selected related entity, weighted by co-occurrence. Own and related histories therefore enter through different routes: older own labels lie along the temporal path, while each related entity contributes one label at one hop.

\item \textbf{\textsc{Chain+Hist-3}.} We build a separate temporal path for the target and for each selected related entity, keeping up to three newest labels in every path, including the target's. The first edge of a related path has its co-occurrence weight; all remaining edges, including the own-history path, have weight $1$. This construction represents recent histories from both sources through the same temporal ordering, without unlabeled relays.

\item \textbf{\textsc{Chain+Hist}.} We remove the three-label limit from \textsc{Chain+Hist-3} and retain all visible labels in both the own and related histories. Each history still forms a newest-first path with the same edge-weight rules, placing older observations farther from the prediction row. Comparing these two variants therefore examines the effect of retaining longer histories under the same attachment rules.
\end{enumerate}

\section{Experimental Results}
\label{app:csresults}
To assess the effect of graph construction, Table~\ref{tab:app-cs} reports all variants evaluated on the same filtered base, and the column names follow the constructions defined in Table~\ref{tab:app-graphnames}. There are three emerging patterns in the experiments. First, improvements are generally small and construction-dependent. Thus, no single construction consistently outperforms the others, and some constructions are detrimental. For example, the self-label star reduces AUROC by $2.7$ points on \texttt{avito}/\texttt{user-clicks}. Second, the exceptions are informative based on a specific database. For instance, on \texttt{trial}/\texttt{study-outcome}, incorporating related-entity labels improves AUROC by up to $+1.67$ points, suggesting that this task contains a stronger cross-entity label signal. In contrast, \texttt{stack}/\texttt{user-engagement} is best served by the pure temporal chain, whose one-hop structure effectively imposes a recency constraint. Third, the same constructions produce substantially larger gains on the unfiltered base, ranging from $+2.8$ to $+4.1$ points on the tasks above. Nevertheless, these models still underperform the filtered base without label propagation on almost all tasks, suggesting that filtered features already capture much of the label signal. Table~\ref{tab:app-cs-reg} summarizes regression results.

\begin{table}[ht]
\centering\small 
\setlength{\tabcolsep}{3.8pt}
\caption{Label-graph variants with Optuna tuning on the trained filtered MLP over 10 seeds.}
\label{tab:app-cs}
\scalebox{0.6}{%
\begin{tabular}{lccccccccc}
\toprule
& \textsc{Self}& \textsc{Cross} & \textsc{Cross+Self} & \textsc{Cross+Hist} & \textsc{All}
& \textsc{Self-Chain} & \textsc{Chain+Cross} & \textsc{Chain+Hist-3} & \textsc{Chain+Hist} \\
\midrule
\multicolumn{10}{l}{\emph{RelBench-v1}} \\
\multicolumn{10}{l}{\emph{classification (AUROC$\times$100 $\uparrow$)}} \\
\texttt{amazon}/\texttt{user-churn} & $68.42{\scriptstyle\pm0.05}$ & $68.47{\scriptstyle\pm0.04}$ & $68.69{\scriptstyle\pm0.04}$ & $68.44{\scriptstyle\pm0.04}$ & $68.61{\scriptstyle\pm0.04}$ & $68.47{\scriptstyle\pm0.05}$ & $68.59{\scriptstyle\pm0.04}$ & \underline{$\mathbf{68.77}{\scriptstyle\pm0.05}$} & $\mathbf{68.76}{\scriptstyle\pm0.05}$ \\
\texttt{amazon}/\texttt{item-churn} & $80.93{\scriptstyle\pm0.06}$ & $81.02{\scriptstyle\pm0.06}$ & $81.03{\scriptstyle\pm0.07}$ & $81.07{\scriptstyle\pm0.06}$ & $\mathbf{81.09}{\scriptstyle\pm0.06}$ & $80.76{\scriptstyle\pm0.06}$ & $81.07{\scriptstyle\pm0.06}$ & \underline{$\mathbf{81.18}{\scriptstyle\pm0.05}$} & \underline{$\mathbf{81.18}{\scriptstyle\pm0.05}$} \\
\texttt{avito}/\texttt{user-clicks} & $62.34{\scriptstyle\pm0.85}$ & $65.08{\scriptstyle\pm1.06}$ & $\mathbf{65.10}{\scriptstyle\pm1.06}$ & \underline{$\mathbf{65.44}{\scriptstyle\pm1.04}$} & $64.86{\scriptstyle\pm1.02}$ & $61.91{\scriptstyle\pm0.77}$ & $64.62{\scriptstyle\pm1.09}$ & $65.09{\scriptstyle\pm1.06}$ & $65.09{\scriptstyle\pm1.06}$ \\
\texttt{avito}/\texttt{user-visits} & $66.43{\scriptstyle\pm0.17}$ & $66.40{\scriptstyle\pm0.15}$ & $66.41{\scriptstyle\pm0.18}$ & $66.36{\scriptstyle\pm0.18}$ & \underline{$\mathbf{66.72}{\scriptstyle\pm0.11}$} & $66.40{\scriptstyle\pm0.17}$ & $\mathbf{66.45}{\scriptstyle\pm0.14}$ & $66.37{\scriptstyle\pm0.18}$ & $66.37{\scriptstyle\pm0.18}$ \\
\texttt{hm}/\texttt{user-churn} & $\mathbf{68.94}{\scriptstyle\pm0.06}$ & $68.23{\scriptstyle\pm0.07}$ & \underline{$\mathbf{69.16}{\scriptstyle\pm0.05}$} & $68.37{\scriptstyle\pm0.08}$ & $68.54{\scriptstyle\pm0.06}$ & $68.46{\scriptstyle\pm0.04}$ & $68.35{\scriptstyle\pm0.07}$ & $68.35{\scriptstyle\pm0.07}$ & $68.35{\scriptstyle\pm0.07}$ \\
\texttt{stack}/\texttt{user-badge} & $\mathbf{86.14}{\scriptstyle\pm0.10}$ & $86.10{\scriptstyle\pm0.08}$ & \underline{$\mathbf{86.16}{\scriptstyle\pm0.09}$} & $86.09{\scriptstyle\pm0.09}$ & $86.10{\scriptstyle\pm0.09}$ & $86.07{\scriptstyle\pm0.08}$ & $86.10{\scriptstyle\pm0.08}$ & $86.11{\scriptstyle\pm0.09}$ & $86.11{\scriptstyle\pm0.09}$ \\
\texttt{stack}/\texttt{user-engagement} & $\mathbf{89.82}{\scriptstyle\pm0.10}$ & $88.60{\scriptstyle\pm0.11}$ & $89.02{\scriptstyle\pm0.09}$ & $88.65{\scriptstyle\pm0.08}$ & $89.04{\scriptstyle\pm0.17}$ & \underline{$\mathbf{90.00}{\scriptstyle\pm0.03}$} & $88.62{\scriptstyle\pm0.11}$ & $88.64{\scriptstyle\pm0.10}$ & $88.64{\scriptstyle\pm0.10}$ \\
\texttt{trial}/\texttt{study-outcome} & $66.37{\scriptstyle\pm0.34}$ & $67.49{\scriptstyle\pm0.38}$ & $\mathbf{67.68}{\scriptstyle\pm0.39}$ & \underline{$\mathbf{68.03}{\scriptstyle\pm0.42}$} & \underline{$\mathbf{68.03}{\scriptstyle\pm0.42}$} & $66.37{\scriptstyle\pm0.34}$ & $\mathbf{67.68}{\scriptstyle\pm0.39}$ & $67.49{\scriptstyle\pm0.38}$ & $\mathbf{67.68}{\scriptstyle\pm0.39}$ \\
\midrule
\multicolumn{10}{l}{\emph{regression (MAE $\downarrow$)}} \\
\texttt{amazon}/\texttt{user-ltv} & $\mathbf{14.594}{\scriptstyle\pm0.009}$ & $14.656{\scriptstyle\pm0.013}$ & $14.656{\scriptstyle\pm0.013}$ & $14.656{\scriptstyle\pm0.013}$ & $14.625{\scriptstyle\pm0.011}$ & \underline{$\mathbf{14.587}{\scriptstyle\pm0.009}$} & $14.656{\scriptstyle\pm0.013}$ & $14.656{\scriptstyle\pm0.013}$ & $14.656{\scriptstyle\pm0.013}$ \\
\texttt{amazon}/\texttt{item-ltv} & $\mathbf{48.029}{\scriptstyle\pm0.177}$ & $50.994{\scriptstyle\pm0.168}$ & $50.994{\scriptstyle\pm0.168}$ & $50.994{\scriptstyle\pm0.168}$ & $50.994{\scriptstyle\pm0.168}$ & \underline{$\mathbf{46.932}{\scriptstyle\pm0.111}$} & $50.995{\scriptstyle\pm0.168}$ & $50.995{\scriptstyle\pm0.168}$ & $50.997{\scriptstyle\pm0.167}$ \\
\texttt{avito}/\texttt{ad-ctr} & $\mathbf{0.039}{\scriptstyle\pm0.003}$ & $0.040{\scriptstyle\pm0.003}$ & $0.040{\scriptstyle\pm0.003}$ & $0.040{\scriptstyle\pm0.003}$ & $0.040{\scriptstyle\pm0.003}$ & \underline{$\mathbf{0.038}{\scriptstyle\pm0.001}$} & $0.040{\scriptstyle\pm0.003}$ & $0.040{\scriptstyle\pm0.003}$ & $0.040{\scriptstyle\pm0.003}$ \\
\texttt{hm}/\texttt{item-sales} & $\mathbf{0.049}{\scriptstyle\pm0.000}$ & $0.060{\scriptstyle\pm0.000}$ & $0.060{\scriptstyle\pm0.000}$ & $0.060{\scriptstyle\pm0.000}$ & $0.060{\scriptstyle\pm0.000}$ & \underline{$\mathbf{0.044}{\scriptstyle\pm0.000}$} & $0.057{\scriptstyle\pm0.000}$ & $0.060{\scriptstyle\pm0.000}$ & $0.060{\scriptstyle\pm0.000}$ \\
\texttt{stack}/\texttt{post-votes} & \underline{$\mathbf{0.066}{\scriptstyle\pm0.000}$} & \underline{$\mathbf{0.066}{\scriptstyle\pm0.000}$} & \underline{$\mathbf{0.066}{\scriptstyle\pm0.000}$} & \underline{$\mathbf{0.066}{\scriptstyle\pm0.000}$} & \underline{$\mathbf{0.066}{\scriptstyle\pm0.000}$} & \underline{$\mathbf{0.066}{\scriptstyle\pm0.000}$} & \underline{$\mathbf{0.066}{\scriptstyle\pm0.000}$} & \underline{$\mathbf{0.066}{\scriptstyle\pm0.000}$} & \underline{$\mathbf{0.066}{\scriptstyle\pm0.000}$} \\
\texttt{trial}/\texttt{site-success} & $0.434{\scriptstyle\pm0.005}$ & $0.445{\scriptstyle\pm0.006}$ & $0.443{\scriptstyle\pm0.006}$ & $0.437{\scriptstyle\pm0.005}$ & $0.437{\scriptstyle\pm0.005}$ & $\mathbf{0.432}{\scriptstyle\pm0.004}$ & $0.454{\scriptstyle\pm0.004}$ & \underline{$\mathbf{0.425}{\scriptstyle\pm0.006}$} & \underline{$\mathbf{0.425}{\scriptstyle\pm0.006}$} \\
\texttt{trial}/\texttt{study-adverse} & $44.867{\scriptstyle\pm0.151}$ & $44.822{\scriptstyle\pm0.150}$ & $\mathbf{44.817}{\scriptstyle\pm0.150}$ & \underline{$\mathbf{44.616}{\scriptstyle\pm0.153}$} & \underline{$\mathbf{44.616}{\scriptstyle\pm0.153}$} & $44.867{\scriptstyle\pm0.151}$ & $44.822{\scriptstyle\pm0.150}$ & $44.822{\scriptstyle\pm0.150}$ & $44.822{\scriptstyle\pm0.150}$ \\
\midrule
\multicolumn{10}{l}{\emph{RelBench-v2}} \\
\multicolumn{10}{l}{\emph{classification (AUROC$\times$100 $\uparrow$)}} \\
\texttt{ratebeer}/\texttt{user-churn} & \underline{$\mathbf{89.83}{\scriptstyle\pm0.10}$} & $89.73{\scriptstyle\pm0.11}$ & $89.73{\scriptstyle\pm0.11}$ & $89.72{\scriptstyle\pm0.12}$ & $89.73{\scriptstyle\pm0.13}$ & $\mathbf{89.81}{\scriptstyle\pm0.11}$ & $89.73{\scriptstyle\pm0.11}$ & $89.74{\scriptstyle\pm0.11}$ & $89.74{\scriptstyle\pm0.11}$ \\
\texttt{ratebeer}/\texttt{beer-churn} & $78.97{\scriptstyle\pm0.06}$ & $78.91{\scriptstyle\pm0.07}$ & $78.90{\scriptstyle\pm0.08}$ & $\mathbf{78.98}{\scriptstyle\pm0.06}$ & $\mathbf{78.98}{\scriptstyle\pm0.06}$ & \underline{$\mathbf{79.00}{\scriptstyle\pm0.05}$} & $78.90{\scriptstyle\pm0.07}$ & $78.90{\scriptstyle\pm0.08}$ & $78.88{\scriptstyle\pm0.08}$ \\
\texttt{ratebeer}/\texttt{brewer-dormant} & $\mathbf{79.82}{\scriptstyle\pm0.04}$ & $79.73{\scriptstyle\pm0.04}$ & $79.78{\scriptstyle\pm0.04}$ & $79.71{\scriptstyle\pm0.04}$ & \underline{$\mathbf{79.83}{\scriptstyle\pm0.04}$} & $\mathbf{79.82}{\scriptstyle\pm0.04}$ & $79.81{\scriptstyle\pm0.04}$ & $79.81{\scriptstyle\pm0.04}$ & $79.81{\scriptstyle\pm0.04}$ \\
\texttt{arxiv}/\texttt{paper-citation} & \underline{$\mathbf{81.68}{\scriptstyle\pm0.04}$} & $81.13{\scriptstyle\pm0.06}$ & $81.17{\scriptstyle\pm0.05}$ & $81.20{\scriptstyle\pm0.05}$ & $81.18{\scriptstyle\pm0.05}$ & $\mathbf{81.36}{\scriptstyle\pm0.06}$ & $81.15{\scriptstyle\pm0.05}$ & $81.15{\scriptstyle\pm0.05}$ & $81.15{\scriptstyle\pm0.05}$ \\
\midrule
\multicolumn{10}{l}{\emph{regression (MAE $\downarrow$)}} \\
\texttt{ratebeer}/\texttt{user-count} & $9.900{\scriptstyle\pm0.058}$ & \underline{$\mathbf{9.115}{\scriptstyle\pm0.075}$} & \underline{$\mathbf{9.115}{\scriptstyle\pm0.075}$} & $9.244{\scriptstyle\pm0.093}$ & $9.294{\scriptstyle\pm0.098}$ & $9.582{\scriptstyle\pm0.035}$ & $\mathbf{9.134}{\scriptstyle\pm0.078}$ & $9.145{\scriptstyle\pm0.080}$ & $9.145{\scriptstyle\pm0.080}$ \\
\texttt{arxiv}/\texttt{author-publication} & \underline{$\mathbf{0.489}{\scriptstyle\pm0.001}$} & \underline{$\mathbf{0.489}{\scriptstyle\pm0.002}$} & \underline{$\mathbf{0.489}{\scriptstyle\pm0.002}$} & \underline{$\mathbf{0.489}{\scriptstyle\pm0.001}$} & \underline{$\mathbf{0.489}{\scriptstyle\pm0.002}$} & \underline{$\mathbf{0.489}{\scriptstyle\pm0.002}$} & \underline{$\mathbf{0.489}{\scriptstyle\pm0.002}$} & \underline{$\mathbf{0.489}{\scriptstyle\pm0.002}$} & \underline{$\mathbf{0.489}{\scriptstyle\pm0.002}$} \\
\bottomrule
\end{tabular}}
\end{table}

To summarize the gains across constructions, Table~\ref{tab:app-cs-best} compares the best variant per task with the two MLP stages. The best construction varies across tasks, with no single construction being best on more than two tasks. The resulting mean improvement therefore reflects per-task selection of the best variant and should be interpreted as an oracle-style upper bound. Any single fixed variant achieves a smaller improvement. For regression, selecting meta-graphs on validation lowers MAE on six of seven tasks (Table~\ref{tab:app-cs-reg}), with five tasks selecting \textsc{Self} or \textsc{Self-Chain}. In particular, these own-history graphs give the largest gains, from $0.060$ to $0.044$ on \texttt{hm}/\texttt{item-sales} and $50.99$ to $46.93$ on \texttt{amazon}/\texttt{item-ltv}, while the two \texttt{trial} tasks select related-entity histories with slight improvements. 

On \texttt{stack}/\texttt{post-votes} and \texttt{arxiv}/\texttt{author-publication}, all nine variants remain within $0.14\%$ of the filtered MLP, as validation favors configurations that largely preserve its predictions. Historical labels are available for $90$--$98\%$ and $61$--$71\%$ of validation rows, respectively, and the target's own past labels correlate with its current value (Pearson $0.61$ and $0.59$). However, this historical signal does not translate into lower MAE. The targets are concentrated at low values, with $95.2\%$ of \texttt{post-votes} targets equal to zero and a median of one on \texttt{author-publication}. Since MAE favors conditional medians, mean-based corrections can increase error even when past labels correlate with the target. For example, adding the mean residual from each row's own label history raises validation MAE from $0.060$ to $0.100$ on \texttt{post-votes} and from $0.428$ to $0.494$ on \texttt{author-publication}.
\begin{table}[ht]
\centering\small\setlength{\tabcolsep}{4pt}
\caption{Best meta-graph per task (test AUROC$\times100$); reporting as in Table~\ref{tab:cls-results}.}
\label{tab:app-cs-best}
\scalebox{0.79}{%
\begin{tabular}{llccc}
\toprule
\textbf{Task} & \textbf{Best variant} & \textbf{MLP (target attributes)} & \textbf{MLP (+relational attributes)} & \textbf{C\&S (+historical labels)} \\
\midrule
\texttt{amazon}/\texttt{user-churn} & \textsc{Chain+Hist-3} & $53.65{\scriptstyle\pm0.18}$ & $\mathbf{68.12}{\scriptstyle\pm0.05}$ & \underline{$\mathbf{68.77}{\scriptstyle\pm0.05}$} \\
\texttt{amazon}/\texttt{item-churn} & \textsc{Chain+Hist-3} & $66.60{\scriptstyle\pm0.44}$ & $\mathbf{80.58}{\scriptstyle\pm0.08}$ & \underline{$\mathbf{81.18}{\scriptstyle\pm0.05}$} \\
\texttt{avito}/\texttt{user-clicks} & \textsc{Cross+Hist} & $51.74{\scriptstyle\pm0.51}$ & $\mathbf{65.08}{\scriptstyle\pm1.06}$ & \underline{$\mathbf{65.44}{\scriptstyle\pm1.04}$} \\
\texttt{avito}/\texttt{user-visits} & \textsc{All} & $49.64{\scriptstyle\pm0.01}$ & $\mathbf{66.29}{\scriptstyle\pm0.20}$ & \underline{$\mathbf{66.72}{\scriptstyle\pm0.11}$} \\
\texttt{hm}/\texttt{user-churn} & \textsc{Cross+Self} & $55.25{\scriptstyle\pm0.05}$ & $\mathbf{68.21}{\scriptstyle\pm0.06}$ & \underline{$\mathbf{69.16}{\scriptstyle\pm0.05}$} \\
\texttt{stack}/\texttt{user-badge} & \textsc{Cross+Self} & $51.93{\scriptstyle\pm1.87}$ & $\mathbf{86.04}{\scriptstyle\pm0.08}$ & \underline{$\mathbf{86.16}{\scriptstyle\pm0.09}$} \\
\texttt{stack}/\texttt{user-engagement} & \textsc{Self-Chain} & $53.04{\scriptstyle\pm0.04}$ & $\mathbf{88.53}{\scriptstyle\pm0.12}$ & \underline{$\mathbf{90.00}{\scriptstyle\pm0.03}$} \\
\texttt{trial}/\texttt{study-outcome} & \textsc{Cross+Hist} & $65.77{\scriptstyle\pm0.36}$ & $\mathbf{66.37}{\scriptstyle\pm0.34}$ & \underline{$\mathbf{68.03}{\scriptstyle\pm0.42}$} \\
\midrule
\textbf{Average} & & $55.95$ & $\mathbf{73.65}$ & \underline{$\mathbf{74.43}$} \\
\bottomrule
\end{tabular}}
\end{table}

\begin{table}[ht]
\centering\small\setlength{\tabcolsep}{4pt}
\caption{Regression with validation-selected meta-graphs.}
\label{tab:app-cs-reg}
\scalebox{0.79}{%
\begin{tabular}{llccc}
\toprule
\textbf{Task} & \textbf{Selected construction} & \textbf{MLP (target attributes)} & \textbf{MLP (+relational attributes)} & \textbf{C\&S (+historical labels)} \\
\midrule
\texttt{amazon}/\texttt{user-ltv} & \textsc{Self-Chain} & $16.757{\scriptstyle\pm0.007}$ & $\mathbf{14.656}{\scriptstyle\pm0.013}$ & \underline{$\mathbf{14.587}{\scriptstyle\pm0.009}$} \\
\texttt{amazon}/\texttt{item-ltv} & \textsc{Self-Chain} & $59.958{\scriptstyle\pm0.150}$ & $\mathbf{50.994}{\scriptstyle\pm0.168}$ & \underline{$\mathbf{46.932}{\scriptstyle\pm0.111}$} \\
\texttt{avito}/\texttt{ad-ctr} & \textsc{Self-Chain} & $0.043{\scriptstyle\pm0.000}$ & $\mathbf{0.040}{\scriptstyle\pm0.003}$ & \underline{$\mathbf{0.038}{\scriptstyle\pm0.001}$} \\
\texttt{hm}/\texttt{item-sales} & \textsc{Self-Chain} & $0.076{\scriptstyle\pm0.000}$ & $\mathbf{0.060}{\scriptstyle\pm0.000}$ & \underline{$\mathbf{0.044}{\scriptstyle\pm0.000}$} \\
\texttt{stack}/\texttt{post-votes} & \textsc{Self} & $\mathbf{0.068}{\scriptstyle\pm0.000}$ & \underline{$\mathbf{0.066}{\scriptstyle\pm0.000}$} & \underline{$\mathbf{0.066}{\scriptstyle\pm0.000}$} \\
\texttt{trial}/\texttt{site-success} & \textsc{Chain+Hist-3} & $\mathbf{0.426}{\scriptstyle\pm0.002}$ & $0.437{\scriptstyle\pm0.006}$ & \underline{$\mathbf{0.425}{\scriptstyle\pm0.006}$} \\
\texttt{trial}/\texttt{study-adverse} & \textsc{Cross+Hist} & $48.203{\scriptstyle\pm0.676}$ & $\mathbf{44.867}{\scriptstyle\pm0.151}$ & \underline{$\mathbf{44.616}{\scriptstyle\pm0.153}$} \\
\bottomrule
\end{tabular}}
\end{table}

\section{Extended Results}\label{app:extended}
In this section, we compare against additional RDL models for completeness, and examine structural features, filter depth, runtime, and historical labels as MLP inputs. Table~\ref{tab:labels-as-features} uses an earlier filtered base with only numerical and text columns. All other tables in this appendix use all column types and the same runs as the main tables.

\subsection{Comparisons with More Models}
Table~\ref{tab:extended-baselines} compares our pipeline with RelGNN~\cite{chen2025relgnn} and RDBLearn~\cite{xu2026no}. RelGNN results are obtained from~\cite{chen2025relgnn}, and RDBLearn results come from~\cite{xu2026no} for RelBench-v1 and~\cite{xu2026parameter} for RelBench-v2.
\begin{table}[ht]
\centering\small\setlength{\tabcolsep}{4pt}
{
\caption{Extended results with more advanced models.}
\label{tab:extended-baselines}
\scalebox{0.76}{%
\begin{tabular}{lcccccc}
\toprule
\textbf{Task} & \makecell{\textbf{MLP}\\\textbf{(target attributes)}} & \makecell{\textbf{MLP}\\\textbf{(+relational attributes)}} & \makecell{\textbf{C\&S}\\\textbf{(+historical labels)}} & \textbf{HeteroGNN} & \textbf{RelGNN} & \textbf{RDBLearn} \\
\midrule
\multicolumn{7}{l}{\emph{RelBench-v1}} \\
\multicolumn{7}{l}{\emph{classification (AUROC$\times$100 $\uparrow$)}} \\
\texttt{amazon}/\texttt{user-churn} & $53.65{\scriptstyle\pm0.18}$ & $68.12{\scriptstyle\pm0.05}$ & $68.69{\scriptstyle\pm0.04}$ & $\mathbf{70.42}{\scriptstyle\pm0.05}$ & \underline{$\mathbf{70.99}$} & $67.57$ \\
\texttt{amazon}/\texttt{item-churn} & $66.60{\scriptstyle\pm0.44}$ & $80.58{\scriptstyle\pm0.08}$ & $81.03{\scriptstyle\pm0.07}$ & \underline{$\mathbf{82.81}{\scriptstyle\pm0.03}$} & $\mathbf{82.64}$ & $82.07$ \\
\texttt{avito}/\texttt{user-clicks} & $51.74{\scriptstyle\pm0.51}$ & $65.08{\scriptstyle\pm1.06}$ & $65.10{\scriptstyle\pm1.06}$ & $65.90{\scriptstyle\pm1.95}$ & $\mathbf{68.23}$ & \underline{$\mathbf{69.04}$} \\
\texttt{avito}/\texttt{user-visits} & $49.64{\scriptstyle\pm0.01}$ & $\mathbf{66.29}{\scriptstyle\pm0.20}$ & \underline{$\mathbf{66.41}{\scriptstyle\pm0.18}$} & $66.20{\scriptstyle\pm0.10}$ & $66.18$ & $65.49$ \\
\texttt{hm}/\texttt{user-churn} & $55.25{\scriptstyle\pm0.05}$ & $68.21{\scriptstyle\pm0.06}$ & $69.16{\scriptstyle\pm0.05}$ & $\mathbf{69.88}{\scriptstyle\pm0.21}$ & \underline{$\mathbf{70.93}$} & $68.05$ \\
\texttt{stack}/\texttt{user-badge} & $51.93{\scriptstyle\pm1.87}$ & $86.04{\scriptstyle\pm0.08}$ & $86.16{\scriptstyle\pm0.09}$ & $\mathbf{88.86}{\scriptstyle\pm0.08}$ & \underline{$\mathbf{88.98}$} & $85.26$ \\
\texttt{stack}/\texttt{user-engagement} & $53.04{\scriptstyle\pm0.04}$ & $88.53{\scriptstyle\pm0.12}$ & $89.02{\scriptstyle\pm0.09}$ & $\mathbf{90.59}{\scriptstyle\pm0.09}$ & \underline{$\mathbf{90.75}$} & $89.39$ \\
\texttt{trial}/\texttt{study-outcome} & $65.77{\scriptstyle\pm0.36}$ & $66.37{\scriptstyle\pm0.34}$ & $67.68{\scriptstyle\pm0.39}$ & $68.60{\scriptstyle\pm1.01}$ & $\mathbf{71.24}$ & \underline{$\mathbf{71.58}$} \\
\midrule
\textbf{Average} & $55.95$ & $73.65$ & $74.16$ & $\mathbf{75.41}$ & \underline{$\mathbf{76.24}$} & $74.81$ \\
\midrule
\multicolumn{7}{l}{\emph{regression (MAE $\downarrow$)}} \\
\texttt{amazon}/\texttt{user-ltv} & $16.757{\scriptstyle\pm0.007}$ & $14.656{\scriptstyle\pm0.013}$ & $14.656{\scriptstyle\pm0.013}$ & $\mathbf{14.313}{\scriptstyle\pm0.013}$ & \underline{$\mathbf{14.230}$} & $14.540$ \\
\texttt{amazon}/\texttt{item-ltv} & $59.958{\scriptstyle\pm0.150}$ & $50.994{\scriptstyle\pm0.168}$ & $50.994{\scriptstyle\pm0.168}$ & $50.053{\scriptstyle\pm0.163}$ & $\mathbf{48.767}$ & \underline{$\mathbf{48.559}$} \\
\texttt{avito}/\texttt{ad-ctr} & $0.043{\scriptstyle\pm0.000}$ & $0.040{\scriptstyle\pm0.003}$ & $0.040{\scriptstyle\pm0.003}$ & $0.041{\scriptstyle\pm0.001}$ & $\mathbf{0.037}$ & \underline{$\mathbf{0.034}$} \\
\texttt{hm}/\texttt{item-sales} & $0.076{\scriptstyle\pm0.000}$ & $0.060{\scriptstyle\pm0.000}$ & $0.060{\scriptstyle\pm0.000}$ & $\mathbf{0.056}{\scriptstyle\pm0.000}$ & \underline{$\mathbf{0.054}$} & $0.064$ \\
\texttt{stack}/\texttt{post-votes} & $0.068{\scriptstyle\pm0.000}$ & $\mathbf{0.066}{\scriptstyle\pm0.000}$ & $\mathbf{0.066}{\scriptstyle\pm0.000}$ & \underline{$\mathbf{0.065}{\scriptstyle\pm0.000}$} & \underline{$\mathbf{0.065}$} & $0.068$ \\
\texttt{trial}/\texttt{site-success} & $0.426{\scriptstyle\pm0.002}$ & $0.437{\scriptstyle\pm0.006}$ & $0.443{\scriptstyle\pm0.006}$ & $\mathbf{0.400}{\scriptstyle\pm0.020}$ & \underline{$\mathbf{0.301}$} & $0.424$ \\
\texttt{trial}/\texttt{study-adverse} & $48.203{\scriptstyle\pm0.676}$ & $44.867{\scriptstyle\pm0.151}$ & $44.817{\scriptstyle\pm0.150}$ & $44.473{\scriptstyle\pm0.209}$ & $\mathbf{44.461}$ & \underline{$\mathbf{43.913}$} \\
\midrule
\multicolumn{7}{l}{\emph{RelBench-v2}} \\
\multicolumn{7}{l}{\emph{classification (AUROC$\times$100 $\uparrow$)}} \\
\texttt{ratebeer}/\texttt{user-churn} & $63.98{\scriptstyle\pm0.00}$ & $89.73{\scriptstyle\pm0.12}$ & $89.73{\scriptstyle\pm0.11}$ & $\mathbf{94.27}{\scriptstyle\pm0.22}$ & - & \underline{$\mathbf{97.84}$} \\
\texttt{ratebeer}/\texttt{beer-churn} & $62.40{\scriptstyle\pm0.84}$ & $\mathbf{78.93}{\scriptstyle\pm0.07}$ & $78.90{\scriptstyle\pm0.08}$ & $78.67{\scriptstyle\pm0.60}$ & - & \underline{$\mathbf{83.56}$} \\
\texttt{ratebeer}/\texttt{brewer-dormant} & $63.67{\scriptstyle\pm0.32}$ & $79.73{\scriptstyle\pm0.04}$ & $79.78{\scriptstyle\pm0.04}$ & $\mathbf{80.51}{\scriptstyle\pm0.17}$ & - & \underline{$\mathbf{81.18}$} \\
\texttt{arxiv}/\texttt{paper-citation} & $73.65{\scriptstyle\pm0.22}$ & $81.11{\scriptstyle\pm0.05}$ & $81.17{\scriptstyle\pm0.05}$ & \underline{$\mathbf{82.50}{\scriptstyle\pm0.04}$} & - & $\mathbf{82.37}$ \\
\midrule
\textbf{Average} & $65.93$ & $82.38$ & $82.40$ & $\mathbf{83.99}$ & - & \underline{$\mathbf{86.24}$} \\
\midrule
\multicolumn{7}{l}{\emph{regression (MAE $\downarrow$)}} \\
\texttt{ratebeer}/\texttt{user-count} & $14.453{\scriptstyle\pm0.006}$ & $9.115{\scriptstyle\pm0.075}$ & $9.115{\scriptstyle\pm0.075}$ & $\mathbf{7.374}{\scriptstyle\pm0.102}$ & - & \underline{$\mathbf{6.870}$} \\
\texttt{arxiv}/\texttt{author-publication} & $0.577{\scriptstyle\pm0.001}$ & $\mathbf{0.489}{\scriptstyle\pm0.002}$ & $\mathbf{0.489}{\scriptstyle\pm0.002}$ & $0.513{\scriptstyle\pm0.008}$ & - & \underline{$\mathbf{0.459}$} \\
\bottomrule
\end{tabular}}}
\end{table}

Our goal is to measure how much predictive signal a simple baseline captures without learned relational aggregation. While these models outperform our pipeline on several tasks, we believe the comparison helps contextualize the performance of our simple baseline and its role in assessing the need for architectural complexity in RDL. Both RelGNN and RDBLearn report only average performance across multiple random seeds. RelGNN~\cite{chen2025relgnn} predates RelBench-v2 and reports no results on it.
\subsection{Results without Structural Columns}
To isolate the contribution of structural information, we compare filtered representations with and without the degree and age features of Eq.~\eqref{eq:structcols}. Both features remain disabled in the $K=0$ baseline. The two rightmost columns of Table~\ref{tab:results-struct} repeat the main results. These results show that our core methodology remains unchanged: relational attributes provide substantial improvements over the $K=0$ baselines, and \emph{C\&S} can still improve the performance by exploiting label propagation.
\begin{table}[ht]
\centering\small\setlength{\tabcolsep}{4pt}
{
\caption{Ablation study on structural columns (degree and age).}
\label{tab:results-struct}
\scalebox{0.72}{%
\begin{tabular}{lc|cc|cc}
\toprule
 & & \multicolumn{2}{c|}{\textbf{Without structural columns}} & \multicolumn{2}{c}{\textbf{With structural columns}} \\
\cmidrule(lr){3-4}\cmidrule(lr){5-6}
\textbf{Task} & \makecell{\textbf{MLP}\\\textbf{(target attributes)}} & \makecell{\textbf{MLP}\\\textbf{(+relational attributes)}} & \makecell{\textbf{C\&S}\\\textbf{(+historical labels)}} & \makecell{\textbf{MLP}\\\textbf{(+relational attributes)}} & \makecell{\textbf{C\&S}\\\textbf{(+historical labels)}} \\
\midrule
\multicolumn{6}{l}{\emph{RelBench-v1}} \\
\multicolumn{6}{l}{\emph{classification (AUROC$\times$100 $\uparrow$)}} \\
\texttt{amazon}/\texttt{user-churn} & $53.65{\scriptstyle\pm0.18}$ & $65.82{\scriptstyle\pm0.10}$ & $67.35{\scriptstyle\pm0.04}$ & $\mathbf{68.12}{\scriptstyle\pm0.05}$ & \underline{$\mathbf{68.69}{\scriptstyle\pm0.04}$} \\
\texttt{amazon}/\texttt{item-churn} & $66.60{\scriptstyle\pm0.44}$ & $75.91{\scriptstyle\pm0.07}$ & $78.07{\scriptstyle\pm0.03}$ & $\mathbf{80.58}{\scriptstyle\pm0.08}$ & \underline{$\mathbf{81.03}{\scriptstyle\pm0.07}$} \\
\texttt{avito}/\texttt{user-clicks} & $51.74{\scriptstyle\pm0.51}$ & \underline{$\mathbf{66.26}{\scriptstyle\pm0.50}$} & \underline{$\mathbf{66.26}{\scriptstyle\pm0.50}$} & $65.08{\scriptstyle\pm1.06}$ & $\mathbf{65.10}{\scriptstyle\pm1.06}$ \\
\texttt{avito}/\texttt{user-visits} & $49.64{\scriptstyle\pm0.01}$ & $66.30{\scriptstyle\pm0.07}$ & \underline{$\mathbf{66.52}{\scriptstyle\pm0.06}$} & $66.29{\scriptstyle\pm0.20}$ & $\mathbf{66.41}{\scriptstyle\pm0.18}$ \\
\texttt{hm}/\texttt{user-churn} & $55.25{\scriptstyle\pm0.05}$ & $68.05{\scriptstyle\pm0.02}$ & $\mathbf{69.08}{\scriptstyle\pm0.06}$ & $68.21{\scriptstyle\pm0.06}$ & \underline{$\mathbf{69.16}{\scriptstyle\pm0.05}$} \\
\texttt{stack}/\texttt{user-badge} & $51.93{\scriptstyle\pm1.87}$ & $84.03{\scriptstyle\pm0.05}$ & $84.94{\scriptstyle\pm0.09}$ & $\mathbf{86.04}{\scriptstyle\pm0.08}$ & \underline{$\mathbf{86.16}{\scriptstyle\pm0.09}$} \\
\texttt{stack}/\texttt{user-engagement} & $53.04{\scriptstyle\pm0.04}$ & $86.56{\scriptstyle\pm0.12}$ & $87.97{\scriptstyle\pm0.10}$ & $\mathbf{88.53}{\scriptstyle\pm0.12}$ & \underline{$\mathbf{89.02}{\scriptstyle\pm0.09}$} \\
\texttt{trial}/\texttt{study-outcome} & $65.77{\scriptstyle\pm0.36}$ & $\mathbf{68.25}{\scriptstyle\pm0.28}$ & \underline{$\mathbf{68.96}{\scriptstyle\pm0.24}$} & $66.37{\scriptstyle\pm0.34}$ & $67.68{\scriptstyle\pm0.39}$ \\
\midrule
\textbf{Average} & $55.95$ & $72.65$ & $73.64$ & $\mathbf{73.65}$ & \underline{$\mathbf{74.16}$} \\
\midrule
\multicolumn{6}{l}{\emph{regression (MAE $\downarrow$)}} \\
\texttt{amazon}/\texttt{user-ltv} & $16.757{\scriptstyle\pm0.007}$ & $\mathbf{15.372}{\scriptstyle\pm0.043}$ & $\mathbf{15.372}{\scriptstyle\pm0.043}$ & \underline{$\mathbf{14.656}{\scriptstyle\pm0.013}$} & \underline{$\mathbf{14.656}{\scriptstyle\pm0.013}$} \\
\texttt{amazon}/\texttt{item-ltv} & $59.958{\scriptstyle\pm0.150}$ & $\mathbf{55.827}{\scriptstyle\pm0.314}$ & $\mathbf{55.827}{\scriptstyle\pm0.314}$ & \underline{$\mathbf{50.994}{\scriptstyle\pm0.168}$} & \underline{$\mathbf{50.994}{\scriptstyle\pm0.168}$} \\
\texttt{avito}/\texttt{ad-ctr} & $0.043{\scriptstyle\pm0.000}$ & $\mathbf{0.040}{\scriptstyle\pm0.001}$ & \underline{$\mathbf{0.039}{\scriptstyle\pm0.000}$} & $\mathbf{0.040}{\scriptstyle\pm0.003}$ & $\mathbf{0.040}{\scriptstyle\pm0.003}$ \\
\texttt{hm}/\texttt{item-sales} & $\mathbf{0.076}{\scriptstyle\pm0.000}$ & $\mathbf{0.076}{\scriptstyle\pm0.000}$ & $\mathbf{0.076}{\scriptstyle\pm0.000}$ & \underline{$\mathbf{0.060}{\scriptstyle\pm0.000}$} & \underline{$\mathbf{0.060}{\scriptstyle\pm0.000}$} \\
\texttt{stack}/\texttt{post-votes} & $\mathbf{0.068}{\scriptstyle\pm0.000}$ & \underline{$\mathbf{0.066}{\scriptstyle\pm0.000}$} & \underline{$\mathbf{0.066}{\scriptstyle\pm0.000}$} & \underline{$\mathbf{0.066}{\scriptstyle\pm0.000}$} & \underline{$\mathbf{0.066}{\scriptstyle\pm0.000}$} \\
\texttt{trial}/\texttt{site-success} & $\mathbf{0.426}{\scriptstyle\pm0.002}$ & \underline{$\mathbf{0.419}{\scriptstyle\pm0.001}$} & $0.427{\scriptstyle\pm0.001}$ & $0.437{\scriptstyle\pm0.006}$ & $0.443{\scriptstyle\pm0.006}$ \\
\texttt{trial}/\texttt{study-adverse} & $48.203{\scriptstyle\pm0.676}$ & $\mathbf{43.781}{\scriptstyle\pm0.239}$ & \underline{$\mathbf{43.738}{\scriptstyle\pm0.238}$} & $44.867{\scriptstyle\pm0.151}$ & $44.817{\scriptstyle\pm0.150}$ \\
\midrule
\multicolumn{6}{l}{\emph{RelBench-v2}} \\
\multicolumn{6}{l}{\emph{classification (AUROC$\times$100 $\uparrow$)}} \\
\texttt{ratebeer}/\texttt{user-churn} & $63.98{\scriptstyle\pm0.00}$ & $89.42{\scriptstyle\pm0.15}$ & $\mathbf{89.49}{\scriptstyle\pm0.13}$ & \underline{$\mathbf{89.73}{\scriptstyle\pm0.12}$} & \underline{$\mathbf{89.73}{\scriptstyle\pm0.11}$} \\
\texttt{ratebeer}/\texttt{beer-churn} & $62.40{\scriptstyle\pm0.84}$ & $76.84{\scriptstyle\pm0.17}$ & $76.81{\scriptstyle\pm0.17}$ & \underline{$\mathbf{78.93}{\scriptstyle\pm0.07}$} & $\mathbf{78.90}{\scriptstyle\pm0.08}$ \\
\texttt{ratebeer}/\texttt{brewer-dormant} & $63.67{\scriptstyle\pm0.32}$ & $77.96{\scriptstyle\pm0.12}$ & $77.95{\scriptstyle\pm0.13}$ & $\mathbf{79.73}{\scriptstyle\pm0.04}$ & \underline{$\mathbf{79.78}{\scriptstyle\pm0.04}$} \\
\texttt{arxiv}/\texttt{paper-citation} & $73.65{\scriptstyle\pm0.22}$ & $79.84{\scriptstyle\pm0.06}$ & $80.64{\scriptstyle\pm0.02}$ & $\mathbf{81.11}{\scriptstyle\pm0.05}$ & \underline{$\mathbf{81.17}{\scriptstyle\pm0.05}$} \\
\midrule
\textbf{Average} & $65.93$ & $81.02$ & $81.22$ & $\mathbf{82.38}$ & \underline{$\mathbf{82.40}$} \\
\midrule
\multicolumn{6}{l}{\emph{regression (MAE $\downarrow$)}} \\
\texttt{ratebeer}/\texttt{user-count} & $14.453{\scriptstyle\pm0.006}$ & $\mathbf{13.270}{\scriptstyle\pm0.323}$ & $\mathbf{13.270}{\scriptstyle\pm0.323}$ & \underline{$\mathbf{9.115}{\scriptstyle\pm0.075}$} & \underline{$\mathbf{9.115}{\scriptstyle\pm0.075}$} \\
\texttt{arxiv}/\texttt{author-publication} & $0.577{\scriptstyle\pm0.001}$ & $0.532{\scriptstyle\pm0.001}$ & $\mathbf{0.526}{\scriptstyle\pm0.001}$ & \underline{$\mathbf{0.489}{\scriptstyle\pm0.002}$} & \underline{$\mathbf{0.489}{\scriptstyle\pm0.002}$} \\
\bottomrule
\end{tabular}}}
\end{table}

\subsection{Filtering Hyperparameters and Runtime}
In this section, we provide further ablation studies on the choice of filtering $K$ and the runtime of our experiments.
\subsubsection{Filtering Choices}
\input{figs/avito_filtering}
We use $K=2$ to match the reference HeteroGNN's two-hop aggregation, then test $K\in\{2,4,6,8,10\}$ on \texttt{rel-avito}. Figure~\ref{fig:avito-filtering} shows that $K=2$ performs best on most tasks, while deeper propagation increases filtering cost without consistent gains.
Similarly, on the ten classification tasks from other databases with results at all tested depths (Figure~\ref{fig:avito-filtering}), $K=2$ is best or tied on five. Deeper filters gain at most $0.21$ AUROC points (\texttt{amazon}/\texttt{user-churn}) but lose up to $7.1$ (\texttt{trial}/\texttt{study-outcome}). Meanwhile, from $K=2$ to $K=8$, filtering takes about $4\times$ longer on \texttt{rel-arxiv} and $630\times$ on \texttt{rel-trial}.
\subsubsection{Runtime}
Table~\ref{tab:app-parallel-filter} compares filtering and training times.
Filtering remains a major computational cost even though the prediction head is small. In particular, for target-row indices $T$, we need only $(\bm{H}_t)_T=(\bm{S}_t^K\bar{\bm{X}}_t)_T$. At $K=2$, associativity gives
\begin{equation}
(\bm{H}_t)_T
=\bigl((\bm{S}_t)_T\bm{S}_t\bigr)\bar{\bm{X}}_t
=(\bm{S}_t)_T\bigl(\bm{S}_t\bar{\bm{X}}_t\bigr).
\end{equation}
The final output contains only target rows, but other visible nodes still contribute through intermediate propagation. As $K$ increases, expanding neighborhoods can make the row-restricted computation more expensive than full propagation, so the implementation switches when the latter is cheaper. The larger neighborhoods can therefore cause a sharp increase in filtering time. For example, on \texttt{rel-avito}, filtering rises from about $5.5$ minutes at $K=2$ to $60$--$77$ minutes at $K=4$, then grows roughly linearly with $K$ (Figure~\ref{fig:avito-filtering}).
\subsection{Ablation Study with Labels as Lagged Features}
To test whether historical labels can replace \emph{C\&S} when used as MLP inputs, we summarize each entity's visible history with twelve features. For prediction row $(v,t)$, let
\begin{equation}
H_v(t)=\{(t_i,y_i)\}_{i=1}^{m}, \qquad t_1<\cdots<t_m<t,
\end{equation}
where $t_i$ is the label observation time. For $m>0$, five features summarize history availability, size, recency, and label values:
\begin{equation}
\begin{aligned}
f_1&=\mathbf{1}[m>0], \\ 
f_2&=\log(1+m), \\ 
f_3&=\log(1+\Delta_{(0)}),\\
f_4&=y_{(0)}, \\
f_5&=\frac{1}{m}\sum_{i=1}^{m}y_i,
\end{aligned}
\end{equation}
where $y_{(0)}$ is the most recent label and $\Delta_{(0)}$ its age in days. To capture recency at different scales, we add five averages weighted by recency rank,
\begin{equation}
\mathrm{EWMA}_{\alpha}=
\frac{\sum_{r=0}^{m-1}(1-\alpha)^r y_{(r)}}{\sum_{r=0}^{m-1}(1-\alpha)^r},
\end{equation}
where $\alpha\in\{0.1,0.3,0.5,0.7,0.9\}$, and rank $r=0$ denotes the newest label and larger $\alpha$ emphasizes recent observations. Two further averages instead use elapsed time:
\begin{equation}
\mathrm{HL}_{\tau}=
\frac{\sum_{i=1}^{m}e^{-\Delta_i/\tau}y_i}{\sum_{i=1}^{m}e^{-\Delta_i/\tau}},
\end{equation}
where $\Delta_i=t-t_i$, and $\tau\in\{30,180\}\text{ days}$. Rows without visible history use an absence indicator and default summary values. We apply the same summaries to related entities' historical labels to form the cross-entity feature block, admitting only labels observed strictly before $t$. Using these summaries, we compare three variants on the same filtered representation: the base MLP, the MLP augmented with self-history features, and the MLP augmented with both self-history and cross-entity label features. We then apply \emph{C\&S} to the latter model using the same cross-entity graph construction described in Appendix~\ref{app:meta-graphs}. Table~\ref{tab:labels-as-features} presents the results on RelBench-v1.
\begin{table}[ht]
\centering\small\setlength{\tabcolsep}{4pt}
{
\caption{Historical labels as features and with C\&S.}
\label{tab:labels-as-features}
\scalebox{0.95}{%
\begin{tabular}{lcccc}
\toprule
\textbf{Task}
& \makecell{\textbf{MLP}\\\textbf{(+relational attributes)}}
& \textbf{+ self hist.}
& \textbf{+ (self+cross) hist.}
& \textbf{+ C\&S} \\
\midrule
\texttt{amazon}/\texttt{user-churn} & $68.09$ & $68.51$ & $\mathbf{69.05}$ & \underline{$\mathbf{69.30}$} \\
\texttt{amazon}/\texttt{item-churn} & $80.58$ & $80.91$ & $\mathbf{81.12}$ & \underline{$\mathbf{81.41}$} \\
\texttt{avito}/\texttt{user-clicks} & $\mathbf{65.35}$ & $65.05$ & \underline{$\mathbf{65.46}$} & $65.03$ \\
\texttt{avito}/\texttt{user-visits} & $66.11$ & $\mathbf{66.55}$ & \underline{$\mathbf{66.68}$} & \underline{$\mathbf{66.68}$} \\
\texttt{hm}/\texttt{user-churn} & $67.84$ & $\mathbf{69.65}$ & \underline{$\mathbf{69.76}$} & \underline{$\mathbf{69.76}$} \\
\texttt{stack}/\texttt{user-badge} & $86.06$ & $86.26$ & $\mathbf{86.29}$ & \underline{$\mathbf{86.34}$} \\
\texttt{stack}/\texttt{user-engagement} & $88.52$ & $90.03$ & $\mathbf{90.27}$ & \underline{$\mathbf{90.46}$} \\
\texttt{trial}/\texttt{study-outcome} & $64.36$ & n/a & $\mathbf{64.71}$ & \underline{$\mathbf{66.47}$} \\
\bottomrule
\end{tabular}}}
\end{table}

Notably, we observe that self-history features improve performance on tasks where an entity's own past labels are informative, with gains of $+1.81$ on \texttt{hm}/\texttt{user-churn} and $+1.51$ on \texttt{stack}/\texttt{user-engagement}. Beyond these feature-based gains, applying \emph{C\&S} to the label-augmented model provides additional improvements on five of eight tasks, including $+1.76$ AUROC on \texttt{trial}/\texttt{study-outcome}, whose target entities have no self-history by construction. Overall, these results show that label propagation can add information beyond fixed history summaries, although the improvements vary across tasks and \texttt{avito}/\texttt{user-clicks} shows a small degradation.

\subsection{Discussion and Open Questions} 
Overall, our studies show that simple propagation methods allow lightweight models to recover a substantial portion of relational predictive power without learned aggregation. Integrating historical label signals further improves performance on tasks where target entities carry informative past outcomes. Nevertheless, the relational database schema itself does not specify which entity connections support effective label propagation. Since shared records create structural links without guaranteeing correlated labels, establishing a canonical method of graph construction for label propagation remains essential for advancing relational deep learning under realistic temporal constraints.   
\end{document}